%% file: main.tex
\documentclass[10pt,twocolumn,letterpaper]{article}

\usepackage{cvpr}              % To produce the CAMERA-READY version
\input{preamble}

\definecolor{cvprblue}{rgb}{0.21,0.49,0.74}
\usepackage[pagebackref,breaklinks,colorlinks,allcolors=cvprblue]{hyperref}

\def\paperID{} % *** Enter the Paper ID here
\def\confName{CVPR}
\def\confYear{2026}
\newcommand{\boldstartspace}[1]
\title{\methodname: Linear-Time Stateful 3D Reconstruction via Test-Time Training}

\author{
    Haian Jin$^{1,2}$
    \quad Rundi Wu$^{1}$
    \quad Tianyuan Zhang$^{3}$
    \quad Ruiqi Gao$^{1}$ \\
    \quad Jonathan T. Barron$^{1}$
    \quad Noah Snavely$^{1,2}$
    \quad Aleksander Hołyński$^{1}$ \\
    \normalsize{
    $^1$ Google DeepMind \quad
    $^2$ Cornell University \quad
    $^3$ Massachusetts Institute of Technology \quad
    }
}

\begin{document}

\twocolumn[{%
% \vspace{-4em}
\vspace{-3.5em}
\maketitle
\thispagestyle{empty}

\begin{center} 
    \vspace{-2em}
    % \vspace{-1.8em}
    \includegraphics[width=\linewidth]{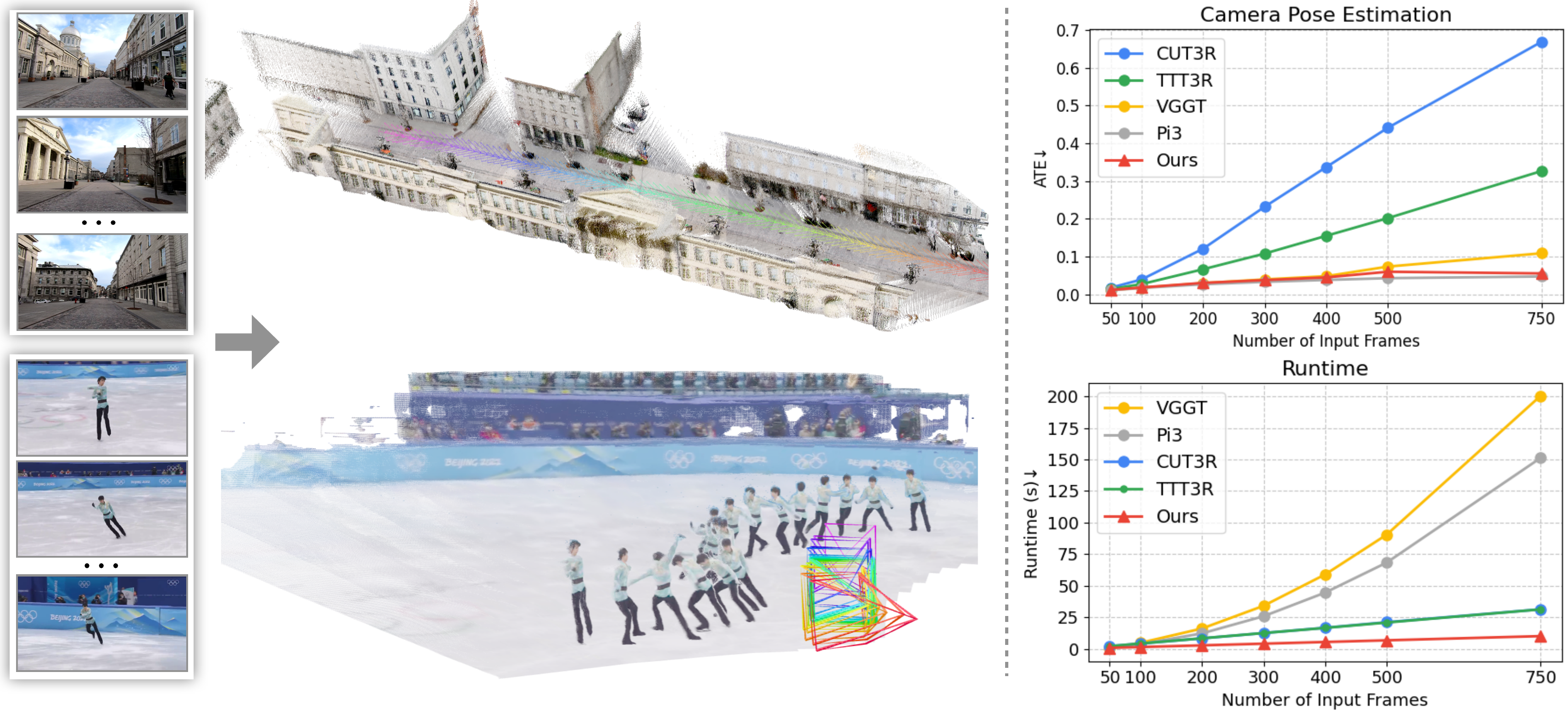}
    % \vspace{-1.5em}
    \captionof{figure}{
    \textbf{\methodname} is an efficient feed-forward 3D reconstruction model whose runtime scales linearly with the number of input views while maintaining or exceeding the reconstruction quality of state-of-the-art quadratic-time systems. 
    \textbf{Left}: Given a long input sequence, \methodname~reconstructs image depths, dense 3D point clouds, and camera trajectory in a single forward pass. 
    \textbf{Right}: 
    Compared to quadratic-time models (VGGT and $\pi^3$), {\methodname} matches or surpasses their prediction accuracy (lower ATE, top) while scaling linearly in runtime (bottom). At 750 frames, our method runs in under 10 seconds, over $20\times$ faster than VGGT. 
    }%
    \label{fig:teaser-fig}%
\end{center} % Matched with \begin{center}
}]

\begin{abstract}
Feed-forward transformer models have driven rapid progress in 3D vision, but state-of-the-art methods such as VGGT and $\pi^3$ have a computational cost that scales quadratically with the number of input images, making them inefficient when applied to large image collections. 
Sequential-reconstruction approaches reduce this cost but sacrifice reconstruction quality. We introduce \textbf{\methodname}, a stateful feed-forward model that achieves linear-time, bidirectional 3D reconstruction while matching or surpassing the accuracy of quadratic-time methods. {\methodname} employs test-time training layers to zip an entire image collection into a compact hidden scene state in a single forward pass, enabling reconstruction of over 700 frames in under 10 seconds on a single H100 GPU—more than \textbf{$20\times$} faster than SOTA methods such as VGGT. Moreover, we demonstrate the benefits of having a stateful representation in real-time scene state querying and its extension to sequential streaming reconstruction.
Project page:
\href{https://haian-jin.github.io/ZipMap}{https://haian-jin.github.io/ZipMap}

\end{abstract}

\section{Introduction}
A long-standing goal in computer vision is reconstructing real-world 3D spaces from images or videos. In recent years, deep learning approaches have become highly effective, with state-of-the-art feed-forward systems like VGGT~\cite{wang2025vggt} %and $\pi^3$~\cite{wang2025pi3}
achieving impressive results. These systems, however, are markedly inefficient for long sequence input, as they rely on expensive global attention mechanisms to establish geometric consistency. As the number of input images to the system increases, the reconstruction time of global attention scales quadratically, making these methods computationally prohibitive to run at scale. Methods like CUT3R~\cite{wang2025cut3r}, Point3R~\cite{point3r}, and TTT3R~\cite{chen2025ttt3r} address this through sequential modeling or local partitioning, but these strategies often reduce reconstruction quality. 

In this work, we introduce \methodname, an efficient, feed-forward model trained for reconstructing large image sets.
Our model combines architectural principles from prior work in large feed-forward transformers~\cite{wang2024dust3r, wang2025vggt} and Test-Time Training~\cite{sun2024learning, zhang2025test} to yield a bidirectional model whose complexity scales linearly with the number of inputs, allowing it to process large image collections in seconds. Unlike prior work, our method achieves these efficiency gains while matching or exceeding the fidelity of expensive, state-of-the-art quadratic-time systems. 

The key to our approach is the use of Test-Time Training (TTT) layers~\cite{zhang2025test}: rather than require expensive global attention across all tokens, our model compresses the entire image collection into a compact hidden state (i.e., into the ``fast-weights" of an MLP) in a single forward pass. This state aggregation is highly efficient and globally coherent, enabling scalability to massive image collections. This stateful representation comes with additional benefits: it serves as an implicit scene representation that can be queried to produce pixel-aligned geometry and appearance at novel viewpoints in real time, and can be readily extended to perform reconstruction in a sequential streaming fashion.

We demonstrate our model's effectiveness on several large-scale datasets, and demonstrate that it not only matches or surpasses the reconstruction quality of prior state-of-the-art models like VGGT, but is significantly faster and more scalable than prior work. As shown in Fig.~\ref{fig:teaser-fig}, when given long input image sequence, \methodname~can reconstruct over 700 images in under 10 seconds (75FPS), which is over \textbf{$20\times$} faster than SOTA methods such as VGGT while delivering comparable or superior accuracy.

\section{Related Work}

\noindent \textbf{Large-scale Structure-from-Motion.} Large scene reconstruction has traditionally relied on Structure-from-Motion (SfM) pipelines. Seminal work like Building Rome in a Day and other methods ~\cite{snavely2008skeletal,agarwal2009building,frahm2010building,furukawa2010towards} have demonstrated the feasibility of city-scale reconstruction, while COLMAP~\cite{schonberger2016colmap} established the standard for accuracy through incremental registration. 
Although recent global methods like GLOMAP~\cite{pan2024global} improve efficiency, classical SfM methods typically yield sparse outputs, require large images overlap, and involve time-consuming multi-view stereo stages.
In contrast, our approach integrates pose and dense geometry prediction into a unified, rapid feed-forward pass.

% \medskip
\smallskip
\noindent \textbf{Feed-forward 3D Reconstruction Models.} 
Recent learning-based models like DUSt3R~\cite{wang2024dust3r} and MAST3R \cite{leroy2024grounding} have demonstrated that dense 3D geometry can be predicted from an image pair in a single feed-forward pass.
This paradigm has been extended beyond the 2-image setting by Fast3R~\cite{yang2025fast3r}, FLARE~\cite{zhang2025flare}, VGGT~\cite{wang2025vggt}, and recent $\pi^3$~\cite{wang2025pi3}. 
However, existing multi-view methods typically rely on standard self-attention to associate structural and pose information across images, causing the computational cost to scale quadratically with the number of images, and limiting their use to relatively small numbers of input images. 
While several approaches accelerate inference via token merging or sparse attention~\cite{shen2025fastvggt, wang2025sparsevggt}, they still retain quadratic runtime complexity.
In contrast, another line of work achieves linear scaling by processing images sequentially~\cite{wang2025cut3r,wang2024spann3r,point3r,chen2025ttt3r}, often at the expense of reconstruction quality. Our work addresses this bottleneck by replacing self-attention based design with a linearly scaling stateful model, which, unlike other sequential solutions, does not require recurrent processing, making it less prone to error accumulation.

\smallskip
\noindent \textbf{Linear Complexity Sequence Models.}  
Efficiently handling long sequences has motivated the development of linear-complexity architectures, especially modern RNNs like Linear Transformers~\cite{katharopoulos2020transformers}, Mamba~\cite{gu2024mamba}, DeltaNet~\cite{schlag2021linear, yang2024parallelizing}, and RWKV~\cite{peng2023rwkv}.
These models maintain relatively small linear recurrent states for efficient GPU parallelization and are primarily designed for 1D causal sequences (e.g., language). Hence, they are not well-suited to our setting with large in-context inputs (hundreds of images) and bidirectional dependencies. 

Recently, Test-Time Training (TTT) layers~\cite{sun2024learning, zhang2025test} have emerged as a powerful framework for linear-complexity sequence models. 
TTT treats part of the model's parameters as ``fast-weight'' memory~\cite{hinton1987using, schmidhuber1992learning}, updated online via gradient descent to capture in-context information. This expands the design space for both linear and nonlinear recurrent architectures~\cite{behrouz2025titans,behrouz2025atlas}. 
Following this, LaCT~\cite{zhang2025test} updates nonlinear MLP fast weights once per large token chunk, improving hardware efficiency and enabling bidirectional context integration. 
{\methodname} is built on LaCT to overcome the scaling limitations of prior feed-forward reconstruction models. By leveraging TTT’s compression, it also summarizes large image input into a compact and queryable scene representation.

\section{Method }

\begin{figure*}[t]
\centering
\includegraphics[width=\linewidth]{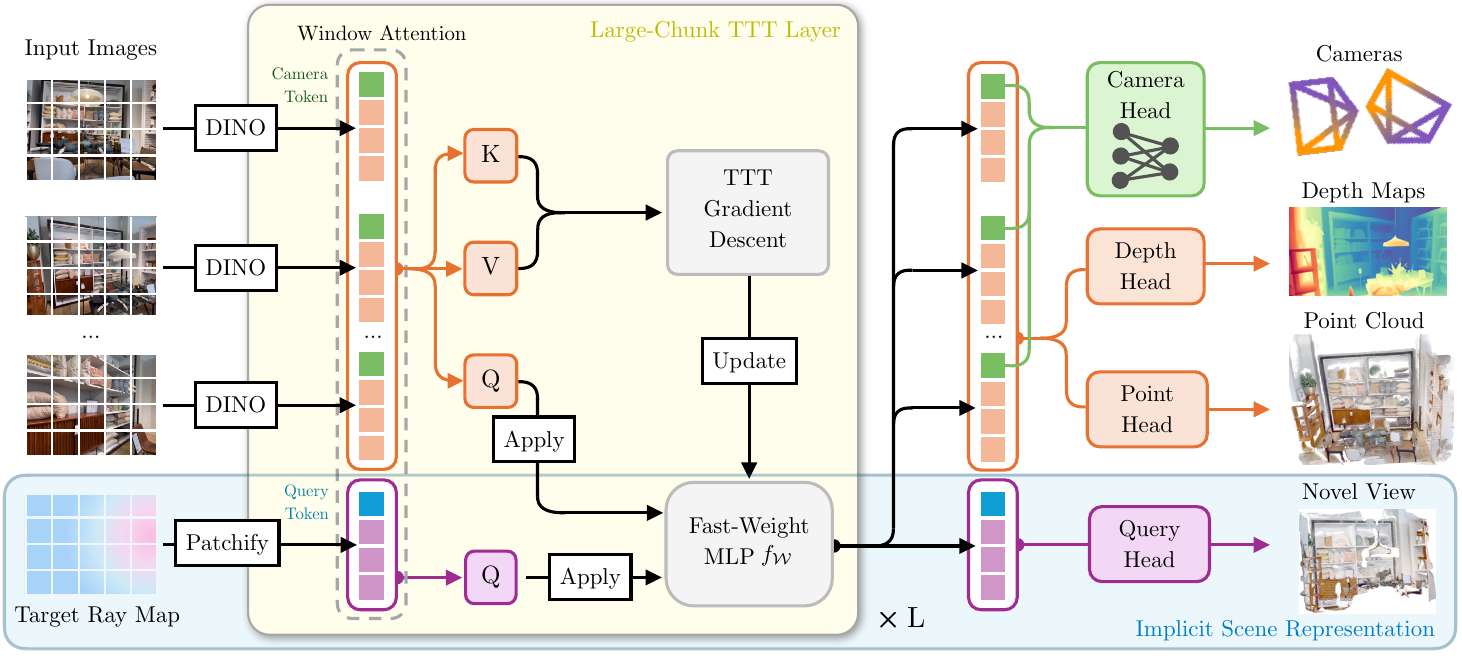}
\vspace{-1.5em}
\caption{
\textbf{Method Overview.}
\methodname\ is a stateful feed-forward model with local window attention and large-chunk TTT layers~\cite{NIPS2017_att, zhang2025test}. Given $N$ input images, a single linear-time pass predicts camera poses, depth maps, and point maps while storing a compact scene representation in TTT fast weights, which can be queried in real time at novel cameras to synthesize new-view point maps.
}
\label{fig:pipeline}
\end{figure*}

An effective 3D foundation model should both reconstruct a 3D scene efficiently and build a queryable, persistent representation. We propose \methodname, a stateful feed-forward model that achieves both in a single pass. Given $N$ input images $\{I_1, \dots, I_N\}$, where $I_{i} \in \mathbb{R}^{H\times W\times 3}$, taken from a video or an unstructured image collection, our model simultaneously achieves:
\begin{enumerate}
    \item \textbf{Efficient 3D Reconstruction:} It supports linear-time, bidirectional reconstruction of camera poses $\{\camera_1, \ldots, \camera_N\}$, depth maps $\{D_1, \dots, D_N\}$, and point clouds $\{\point_1, \ldots, \point_N\}$, all in a single feed-forward pass. %\red{Moreover, the same stateful architecture also supports streaming reconstruction via sequential state updates.}
    \item \textbf{Implicit Scene Representation:} In the same pass, the model automatically adapts its weights to form a queryable implicit scene representation. Given a target camera condition $\camera^t$, it can be queried \emph{in real-time} to produce a colored point-map from that novel viewpoint.
\end{enumerate}

\noindent The key to our model is its efficient design: the runtime of the feed-forward pass scales \emph{linearly} with the number of input images, rather than the quadratic scaling of single-pass models such as VGGT~\cite{wang2025vggt} and the recent $\pi^3$~\cite{wang2025pi3}. 
To achieve linear scaling, instead of using global attention as in prior work~\cite{yang2025fast3r, wang2025vggt, wang2025pi3, zhang2025flare, keetha2026mapanything}, our architecture consists of local window attention and a global large-chunk test-time training(TTT) block~\cite{zhang2025test}. 
Unlike standard attention, which maintains a growing buffer of tokens, TTT compresses the visual context into a fixed-size set of ``fast weights", enabling $\mathcal{O}(N)$ bidirectional reconstruction while yielding a implicit scene state that can be queried from novel viewpoints in constant real time, independent of $N$.

\subsection{Input Tokenization}

Our model can processes two types of inputs: image inputs $\mathcal{I} = \{I_1, \dots, I_N\}$ for scene reconstruction, and a target ray map $\raymap$ , used to query the implicit representation constructed within our model.

We first tokenize each input image $I_i$ using a pretrained DINOv2 encoder~\cite{oquab2023dinov2}. The encoder outputs a 2D spatial feature map, which we flatten into a sequence of patch-level tokens, serving as the primary input for the 3D reconstruction component.
The ray map input $\raymap \in \mathbb{R}^{H \times W \times 9}$ is computed from the target camera's extrinsic and intrinsic parameters. Each 9-dimensional pixel concatenates the ray origin $\vect{r}_o\in\mathbb{R}^3$, direction $\vect{r}_d\in\mathbb{R}^3$, and $\vect{r}_o\times\vect{r}_d$. We patchify $\raymap$ into non-overlapping patches, flatten each patch, and project it with a linear layer into the token embedding dimension.

Additionally, as in \cite{wang2025vggt}, we assign each input image one camera token that will be used to predict its camera information, alongside four register tokens. For the ray-map input, we replace the camera token with a special query token. 
After the tokenization, the image tokens are $\{\vect{x}_i\}_{i=1}^{N}$ with $\vect{x}_i\in\mathbb{R}^{p\times d}$, where $p=HW/P^2+5$ and $P=14$ (patch size). The ray-map tokens are $\{\vect{t}_i\}_{i=1}^{M}$ with $\vect{t}_i\in\mathbb{R}^{p\times d}$.

\subsection{Feature Backbone}
After per-frame tokenization, the model backbone processes all frames jointly to integrate global information to infer the 3D information for the input images, such as camera and depth, 
Prior reconstruction models typically rely on transformers with full attention or alternating global–local attention as the backbone (e.g., VGGT~\cite{wang2025vggt}), incurring quadratic cost in the number of input images. In contrast, we design a linear-cost backbone by replacing all global attention with a large-chunk test-time training (TTT) layer~\cite{zhang2025test}, which compresses all image features into a nonlinear fast-weight function. Our backbone interleaves per-frame local window attention with global TTT layers. Specifically,  it contains $L=24$ identical blocks, each composed of:
\begin{enumerate}
\item \textbf{Local Window Attention} operates on the tokens of each view (image or ray map) independently. 
It uses standard self-attention with rotary positional encoding~\cite{su2024rope} to capture local spatial relationships within each view. 
% It uses standard self-attention within non-overlapping spatial windows, which lets it efficiently process high-resolution feature maps and capture local spatial relationships within each view. 
\item \textbf{Global Large-Chunk TTT Layer} inspired by LaCT~\cite{zhang2025test}, is the key to both our model's linear scaling and its adaptive implicit scene representation. It aggregates global information by updating a nonlinear fast-weight function over all input image tokens. We detail this mechanism below.
\end{enumerate}

\noindent The core of the TTT block is an in-context–adapted fast-weight function, implemented as a SwiGLU-MLP~\cite{shazeer2020glu}:
\begin{equation}
f_\wmatset (\vect{x}) = \wmat_2 \bigg( \operatorname{SiLU} \lft(\wmat_1 \vect{x} \rgt) \circ \lft(\wmat_3 \vect{x} \rgt) \bigg),
\end{equation}
where $\circ$ is elementwise multiplication and $\wmatset=\{\wmat_1, \wmat_2, \wmat_3\}$ are the fast weights.

These fast weights are adapted using a single gradient descent step over tokens from all input views, with a virtual test-time training objective based on key–value reconstruction. Specifically, we project each token  into the corresponding query $\vect{q}_i$, key $\vect{k}_i$, and value $\vect{v}_i$ vector spaces. The key-value pairs from all input image tokens define a virtual objective function:
\begin{align}\label{eq:ttt_loss}
\mathcal{L}\lft(f_\wmatset \lft(\vect{k}_i \rgt), \vect{v}_i \rgt) = -f_\wmatset(\vect{k}_i)^\top \vect{v}_i\,,
\end{align}
which encourages the fast-weight function to memorize the mapping from each key vector to its corresponding value vector. This virtual objective is unrelated to the 3D reconstruction loss; it is optimized once per layer to build an in-context associative memory~\cite{wang2025test}.

To optimize this virtual objective, we first compute the fast-weight gradient $\vect{g}$ of this objective:
\begin{align}
\vect{g} &= \nabla_{\wmatset} \sum_{k=1}^{N \times p} \eta_i \mathcal{L}(f_{\wmatset}(\vect{k}_i), \vect{v}_i) \,,
\end{align}
where $\eta_i$ is the learning rate for each token, as predicted by a simple linear layer that takes as input token. 

Following the Muon~\cite{jordan2024muon} optimizer, we apply the Newton–Schulz orthonormalization procedure to the gradient $\vect{g}$, then update the fast weights followed by L2 normalization to maintain stability:
\begin{align}
\Delta &\leftarrow \operatorname{NewtonSchulz}(\vect{g}) \\
\hat{\wmatset} & \leftarrow \norm{\wmatset} \cdot \frac{\wmatset - \Delta}{\norm{\wmatset - \Delta}} 
\end{align}
These updated fast weights encode global information about the scene. We then apply the updated fast weights $\hat{\wmatset}$ to the input image tokens by passing each token's query $\vect{q}_i$ through the updated fast-weight MLPs: 
\begin{equation}
\vect{o}^\prime_i  = f_{\hat{\wmatset}}(\vect{q}_i),
\end{equation}
Applying the updated fast-weight MLP $f_{\hat{\wmatset}}$ to the input query tokens is analogous to querying all key-value pairs in self-attention, but with linear rather than quadratic complexity in the number of tokens.

These same fast weights can be directly applied to the query $\vect{q}_k^{t}$ of target ray tokens obtained from raymap $\raymap$. This serves a similar role as cross-attending to input image tokens using the target ray tokens. This apply operation has constant runtime per target ray token, independent of the number of input views used to update the fast weights.

Finally, inspired by gated attention~\cite{qiu2025gated, hua2022transformerqualitylineartime}, we apply a gated unit $\operatorname{SiLU} \lft( \mat{W}_g \vect{o}_i \rgt)$ parameterized by weights $\mat{W}_g$ to produce the final output:
\begin{equation}
\vect{o}_i  = \operatorname{RMSNorm}(\vect{o}^\prime_i)\cdot \operatorname{SiLU} \lft( \mat{W}_g \vect{o}^\prime_i \rgt) \,.
\end{equation}

\subsection{Streaming Reconstruction}
\label{sec:sequential_ttt}
The above section performs bidirectional reconstruction by updating each TTT layer once using visual tokens from all input views. 
{\methodname} can also be extended to \emph{streaming reconstruction} by updating its fast weights online, one view at a time. For an image stream $\{I_1, I_2, \ldots\}$, we sequentially update the TTT fast weights $\wmatset^{(t)}$:
\begin{equation}
\wmatset^{(t)} \leftarrow \operatorname{TTTUpdate}\!\left(\wmatset^{(t-1)}; \{\vect{k}_{t,i}, \vect{v}_{t,i}\}_{i=1}^{p}\right),
\end{equation}
using the same virtual key--value objective as in Eq.~\ref{eq:ttt_loss}, but computed only from the visual tokens of the current view. The main paper mainly focuses on the linear time, bidirectional reconstruction. We further evaluate the streaming setting the Appendix~\ref{appendix:streaming_recon}.

\subsection{Prediction Heads}
Our model has four prediction heads. We adopt the same camera head design as VGGT and predict the camera parameters $\camera_i \in \mathbb{R}^{9}$ as a 4D rotation quaternion, 3D translation, and two intrinsics from the input’s camera tokens. We use a DPT-style head~\cite{DPT_Ranftl2021} for the point, depth, and query heads. 

For each input image $i$, the point head predicts a local point map $P_i \in \mathbb{R}^{H \times W \times 3}$ in camera coordinates, similar to $\pi^3$~\cite{wang2025pi3}.
Unlike $\pi^3$, we additionally include a depth head to predict a depth map $D_i \in \mathbb{R}^{H \times W}$ and corresponding confidence map $\Sigma_i \in \mathbb{R}^{H \times W}$.
We find that while either head yields similar quantitative performance, the depth head produces visually smoother results,
and the self-learned confidence map $\Sigma$ helps filter noisy pixels at inference.
Following prior work~\cite{wang2025cut3r, jin2025lvsm,jiang2025rayzer,srt22,sajjadi2022rust}, the query head directly predicts target-view RGB values $I^t \in \mathbb{R}^{H \times W \times 3}$ without an explicit scene representation, and it additionally queries geometry by predicting a target depth map $D^t \in \mathbb{R}^{H \times W}$ with confidence $\Sigma^t \in \mathbb{R}^{H \times W}$.

\newcommand{\subscripttext}[1]{\mathit{#1}}

\subsection{Model Training}
\medskip \noindent \textbf{Training Losses.} We train our model by minimizing the sum of multiple loss functions:
\begin{equation}\label{eq:training_loss}
\mathcal{L}
= \mathcal{L}_\text{point} + \mathcal{L}_\text{depth} + w_c \times \mathcal{L}_\text{cam} \left( + \mathcal{L}_\text{color}^t + \mathcal{L}_{\text{depth}}^{t}\right)
\end{equation}
We follow the open-source implementation of VGGT~\cite{wang2025vggt} and set $w_c = 5$. 
The query losses $\mathcal{L}_\text{color}^t$ and $\mathcal{L}_{\text{depth}}^{t}$ are only enabled during finetuning and are defined w.r.t.\ the ground-truth target image/depth (not the inputs).

\medskip \noindent \textbf{Point loss.} Following~\cite{wang2025pi3,wang2025moge}, we use a scale-invariant local point reconstruction loss:
\begin{align}
    \mathcal{L}_{\text{point}} =\operatorname*{mean}_{i, j} \lft( \frac{ \norm{ \hat{s} \point_{i,j} - \point^*_{i,j} }_1 }{3 z^*_{i,j}} \rgt)\,.
\end{align}
Here, $\point_{i,j}\in\mathbb{R}^3$ is the predicted pixel-aligned point in view $i$ at pixel $j$ (in the local camera coordinates), $\point^*_{i,j}$ is the ground truth, and $z^*_{i,j}$ is the corresponding depth (the $z$-component of $\point^*_{i,j}$). We estimate the global scale $\hat{s}$ via:

\begin{equation}
\label{eq:scale_align}
\hat{s} = \underset{s}{\arg\min} \sum_{i,j} \frac{\norm{s \point_{i,j} - \point^*_{i,j} }_1}{z^*_{i,j}}\,.
\end{equation}
This optimization sub-problem is solved using the ROE solver proposed by~\cite{wang2025moge}.

\medskip \noindent \textbf{Depth loss.} Using that same scale factor $\hat{s}$, we compute our depth loss as:
\begin{equation}
\mathcal{L}_{\text{depth}} = \operatorname*{mean}_i \big( \norm{\Sigma_i \circ \lft(\hat{s} D_i - D^*_i \rgt)}_1 - \alpha \log \Sigma_i \big)\,.
\label{eq:depth_loss}
\end{equation}
This is the $L_1$ loss between the scale-normalized depth prediction $\hat{s} D_i$ and the ground truth depth $D^*_i$ modulated by the predicted uncertainty map $\Sigma_i$, minus a scaled logarithm of $\Sigma_i$ to prevent degenerate solutions and to cause the loss to behave equivalently to the negative log-likelihood of a Laplacian distribution. We set $\alpha=0.2$.

\medskip \noindent \textbf{Camera loss.} We first train with the first image as a reference view, using an $L_1$ loss on camera parameters:
\begin{equation}
\mathcal{L}_{\text{cam}} = \frac{1}{N}\sum_{i=1}^{N} \norm{ \camera^\prime_i - \camera^*_i }\,.
\label{eq:camera_loss}
\end{equation}
where $\camera'_i$ scales the predicted translation by $\hat{s}$.
We then remove the reference view and switch to an affine-invariant camera loss (inspired by $\pi^3$~\cite{wang2025pi3}). We observe that removing the reference view has a limited impact on standard benchmarks, but improves performance on long-sequence inputs.

\medskip \noindent \textbf{Smooth loss.} We additionally apply a normal loss on point maps and a depth gradient loss for local smoothness (Appendix~\ref{appendix:full_training_loss})

\medskip \noindent \textbf{Query loss.} To enable query the implicit scene state,, we additionally include the query
losses $\mathcal{L}_\text{color}^t$ and $\mathcal{L}_{\text{depth}}^{t}$ to fientune the model. We set $\mathcal{L}_\text{color}^t = 10 \times (\text{MSE}+\text{LPIPS})$ between the predicted target RGB and ground truth, and define $\mathcal{L}_{\text{depth}}^{t}$ using Eq.~\eqref{eq:depth_loss} with $(D^t,\Sigma^t)$ and the target view depth ground truth.

\medskip \noindent \textbf{Implementation Details.} Our model uses 24 layers of local window attention interleaved with large-chunk test-time training (TTT) blocks. We re-use VGGT's DINOv2 encoder and initialize our local window attention weights from VGGT's frame-wise attention~\cite{wang2025vggt}. We also initialize a subset of the TTT parameters from VGGT’s global-attention parameters. We set the token dimension to $d = 1024$ and the intermediate dimension of the fast-weight MLP to $2048$, resulting in a state size of $6d^2$ per layer.

We train our model on $64$ H100 GPUs in three stages. First, we train on static datasets with a designated reference view for $80\text{K}$ iterations, using a learning rate of $1\text{e}{-4}$ for the TTT blocks and $1\text{e}{-5}$ for all other modules; this stage takes about $5$ days. Next, we incorporate dynamic datasets and fine-tune for $40\text{K}$ iterations with a uniform learning rate of $1\text{e}{-5}$ for approximately $2.5$ days. Finally, we remove the reference view and train for an additional $60\text{K}$ iterations with a learning rate of $1\text{e}{-5}$. Additional details on it and the fine-tuning procedure used to enable streaming reconstruction and scene-state querying are provided in the Appendix~\ref{appendix:more_imple_details}.

\medskip \noindent \textbf{Training Datasets.} We train our model on a diverse collection of 29 publicly available datasets. Detailed dataset information is provided in Appendix~\ref{appendix:training_datasets}.

\section{Experiments}
\label{sec:experiments}
We evaluate \methodname~across a comprehensive suite of 3D tasks: camera pose, point map, and video/monocular depth estimation. Experiments show that our TTT-based architecture matches or surpasses state-of-the-art quadratic-time models (e.g., VGGT~\cite{wang2025vggt} and $\pi^3$~\cite{wang2025pi3}) while being significantly more compute-efficient.

\begin{figure*}[thb]
\centering
\includegraphics[width=\linewidth]{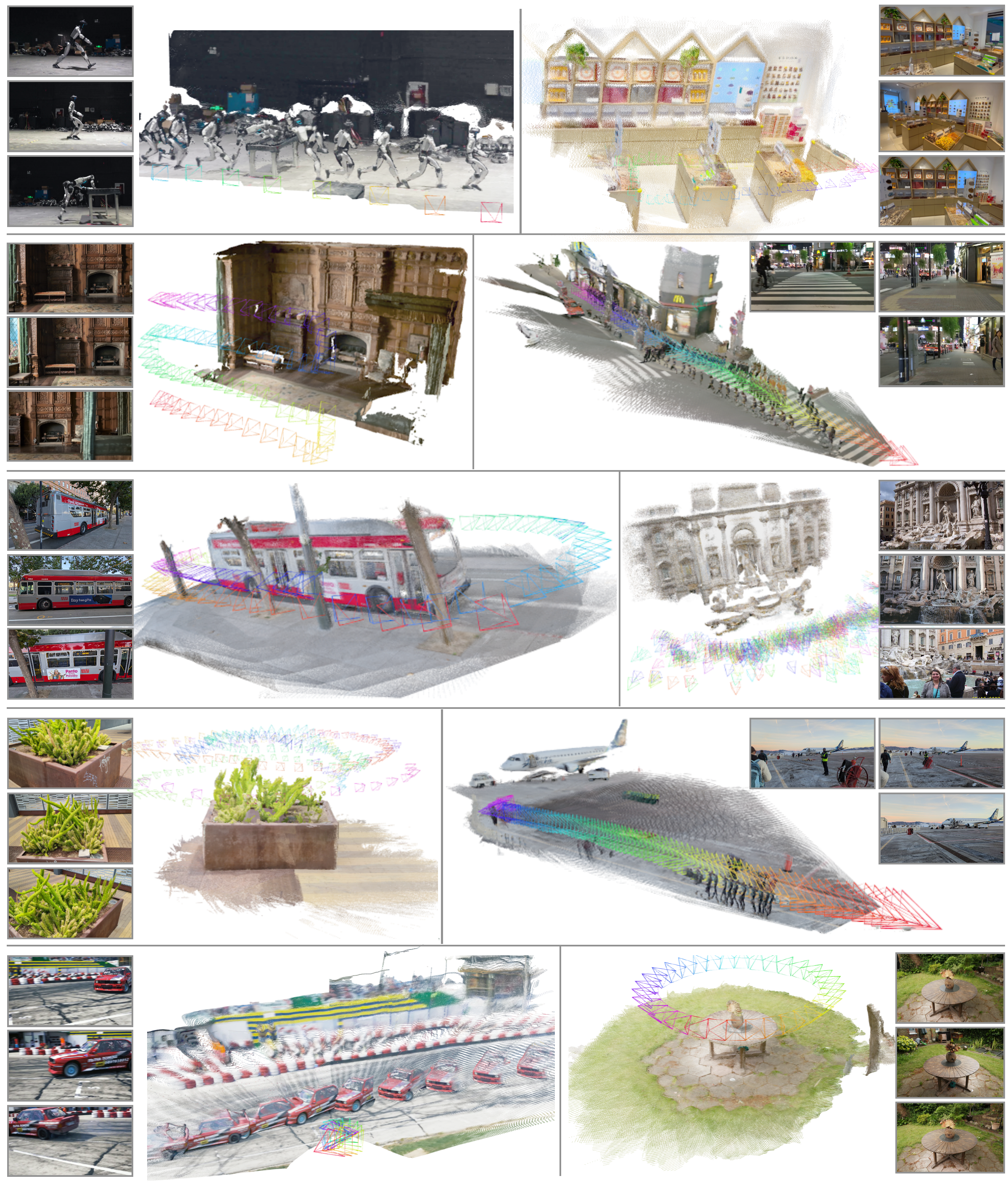}
\caption{\textbf{Example reconstruction results} A sparse subset of input images are shown on the left, and a visualization of the output 3D reconstructions are shown on the right. Note that our method performs well on challenging cases like long sequence inputs, dynamic scenes and internet photo collections.}
\label{fig:reconresults}
\end{figure*}

\subsection{Benchmark Evaluation
}
\label{sec:benchmark_eval}

\input{figures/tables}

\medskip \noindent \textbf{Camera Pose Estimation.} Tables~\ref{tab:camera_re10k_co3d} and~\ref{tab:camera_sintel_tum_scannet} evaluate camera pose accuracy on  RealEstate10K~\cite{zhou2018stereo}, Co3Dv2~\cite{reizenstein21co3d}, Sintel~\cite{butler2012sintel}, TUM Dynamics~\cite{Sturm2012ABF}, and ScanNet~\cite{dai2017scannet}. Our results is comparable to the prior state-of-the-art VGGT~\cite{wang2025vggt} and the recent best method, $\pi^3$~\cite{wang2025pi3}. Crucially, we achieve this accuracy while maintaining linear computational complexity, making our approach notably faster and more scalable than these quadratic-time baselines.

\medskip \noindent \textbf{Point Map Estimation.} Table~\ref{tab:pointmap_7s_nrgbd} and \ref{tab:pointmap_dtu_eth3d} evaluate dense geometry reconstruction on 7-Scenes~\cite{Shotton2013SceneCR}, NRGBD~\cite{Azinovic_2022_CVPR}, DTU~\cite{Jensen2014LargeSM} and ETH3D~\cite{schoeps2017cvpr}. Our model substantially outperforms linear-time baselines such as CUT3R~\cite{wang2025cut3r} and TTT3R~\cite{chen2025ttt3r}, while matching or exceeding the reconstruction quality of state-of-the-art quadratic models including VGGT~\cite{wang2025vggt} and the recent $\pi^3$~\cite{wang2025pi3}. In the Appendix, we provide qualitative comparisons in Figure~\ref{fig:qualitative_comparison}. Figure~\ref{fig:reconresults} further demonstrates that our method recovers coherent 3D structure even in challenging dynamic scenes, where prior approaches often fail.

\medskip \noindent \textbf{Depth Estimation.} We evaluate depth estimation in two contexts: video depth (Table~\ref{tab:videodepth_scale}) and monocular depth (Table~\ref{tab:monodepth} in Appendix). For video depth, we test on Sintel~\cite{butler2012sintel}, Bonn~\cite{Palazzolo2019ReFusion3R}, and KITTI~\cite{Geiger2013IJRR}. Our model consistently outperforms other $\mathcal{O}(N)$ methods and generally exceeds the strong $\mathcal{O}(N^2)$ baseline VGGT. For monocular depth estimation, where frames are evaluated independently, our model remains highly competitive. 
In this setting, we also evaluate on the NYU-v2 dataset~\cite{silberman2012indoor}, where we outperform all baselines considerably, confirming that our method maintains strong single-view geometric priors.

\begin{figure}[thp]
\centering
\includegraphics[width=\linewidth]{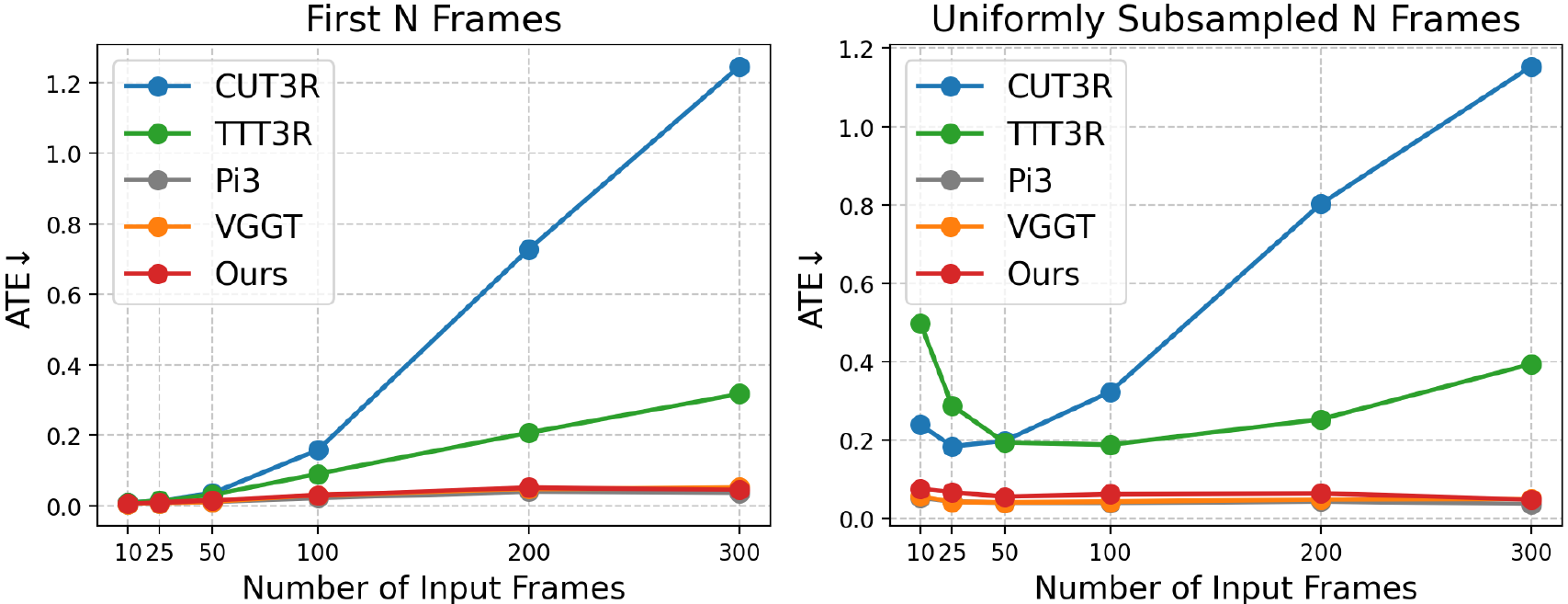}
\caption{
\textbf{Long-sequence camera evaluation on DL3DV.} We evaluate camera pose accuracy (ATE$\downarrow$) on the DL3DV test set~\cite{ling2024dl3dv} under two protocols: \textbf{Left}: increasing \emph{scene scale} by using the first $N$ frames of each sequence; \textbf{Right}: increasing \emph{view density} by uniformly subsampling $N$ frames along a fixed trajectory. Our method maintains low error and matches quadratic-time baselines ($\pi^3$, VGGT) while other linear-time methods (CUT3R, TTT3R) degrade significantly as $N$ grows.
}
\label{fig:long-seq-eval}
\end{figure}

\subsection{Efficiency and Scalability}

As shown in Figure~\ref{fig:teaser-fig}, {\methodname} achieves exceptional reconstruction speed with linear scaling. The bottom plot shows that it reconstructs over 700 frames in 10 seconds on a single H100 GPU, while a quadratic method like VGGT~\cite{wang2025vggt} requires over 200 seconds. We are also about $3\times$ faster than previous linear-time methods like CUT3R and TTT3R, despite their smaller models. This is largely because these baselines reconstruct frames sequentially (one at a time), leading to lower GPU utilization at inference time. Full runtime and evaluation details are provided in Appendix~\ref{appendix:runtime_eval}.

Importantly, this speed does not come at the cost of quality. The top plot in Figure~\ref{fig:teaser-fig} shows that {\methodname} attains a final Absolute Trajectory Error (ATE) on ScanNetV2~\cite{dai2017scannet} comparable to the highly accurate $\pi^3$ and better than VGGT, while substantially outperforming CUT3R and TTT3R. Figure~\ref{fig:long-seq-eval} further analyzes long-sequence behavior on outdoor scenes from the DL3DV test set~\cite{ling2024dl3dv}. We consider two cases: (1) increasing \emph{scene scale} by taking the first $N$ frames, and (2) increasing \emph{view density} by uniformly subsampling $N$ frames over a fixed trajectory. In both settings, {\methodname} matches the accuracy of quadratic-time methods ($\pi^3$, VGGT) while significantly outperforming other linear-time methods (CUT3R, TTT3R), whose errors grow sharply as the number of input frames increases. More results and analysis of long-sequence evaluation in Appendix~\ref{appendix:more_long_eval}.

\subsection{Ablation Studies}
\begin{table}[t]
    \centering
    \caption{
        \textbf{Ablation Study for TTT key components.} We evaluate point-map estimation on ETH3D~\cite{schoeps2017cvpr}. All variants are trained with reduced compute and a smaller data scale.
    }
    \vspace{-1em}
    
    \fontsize{7pt}{8pt}\selectfont
    
    \setlength{\tabcolsep}{3pt}
    \begin{tabular}{l c c c c c c}
        \toprule
        
        % --- HEADER ROW 1 ---
        \multirow{3}{*}{\textbf{Method}} &
        \multicolumn{2}{c}{Acc. $\downarrow$}  &
        \multicolumn{2}{c}{Comp. $\downarrow$} &
        \multicolumn{2}{c}{N.C. $\uparrow$}    \\
        \cmidrule(lr){2-3} \cmidrule(lr){4-5} \cmidrule(lr){6-7}
        
        % --- HEADER ROW 2 ---
        & Mean & Med. & Mean & Med. & Mean & Med. \\
        \midrule
        
        \textbf{Ours} & \textbf{0.337} & \textbf{0.224} & \textbf{0.357} & \textbf{0.217} & \textbf{0.810} & \textbf{0.918} \\
        Ours w/o gated unit & 0.354 & 0.251 & 0.381 & 0.234 & 0.802 & 0.901 \\
        Ours w/o Newton--Schulz & 0.408 & 0.283 & 0.430 & 0.249 & 0.787 & 0.898 \\
        Ours w/ global TTT lr (0.1) & 0.411 & 0.303 & 0.490 & 0.317 & 0.779 & 0.886 \\
        Ours w/ global TTT lr (1.0) & 0.464 & 0.366 & 0.537 & 0.343 & 0.782 & 0.890 \\
        
        \bottomrule
    \end{tabular}
    \label{tab:ttt_abla}
\end{table}

\medskip \noindent \textbf{TTT Components}. As shown in Tab.~\ref{tab:ttt_abla}, Newton--Schulz normalization (Eq.~4) and gated unit (Eq.~7) are crucial; removing either degrades performance. Dynamic per-token learning rate $\eta(x)$ also clearly outperforms fixed global TTT learning rates (0.1 or 1.0).

\medskip \noindent \textbf{Removing the Reference View}. In the final training stage, we remove the explicit reference-view selection, and instead train with the affine-invariant loss proposed in $\pi^3$~\cite{wang2025pi3}. In our setting, removing the reference view  doesn't yields a clear or consistent advantage on the standard benchmarks in Sec.~\ref{sec:benchmark_eval}. However, we observe that it improves accuracy and generalization on long input sequences. Additional details are provided in Appendix~\ref{appendix:remove_ref_Vew}.

\begin{figure}[tb]
    \centering
    \includegraphics[width=\linewidth]{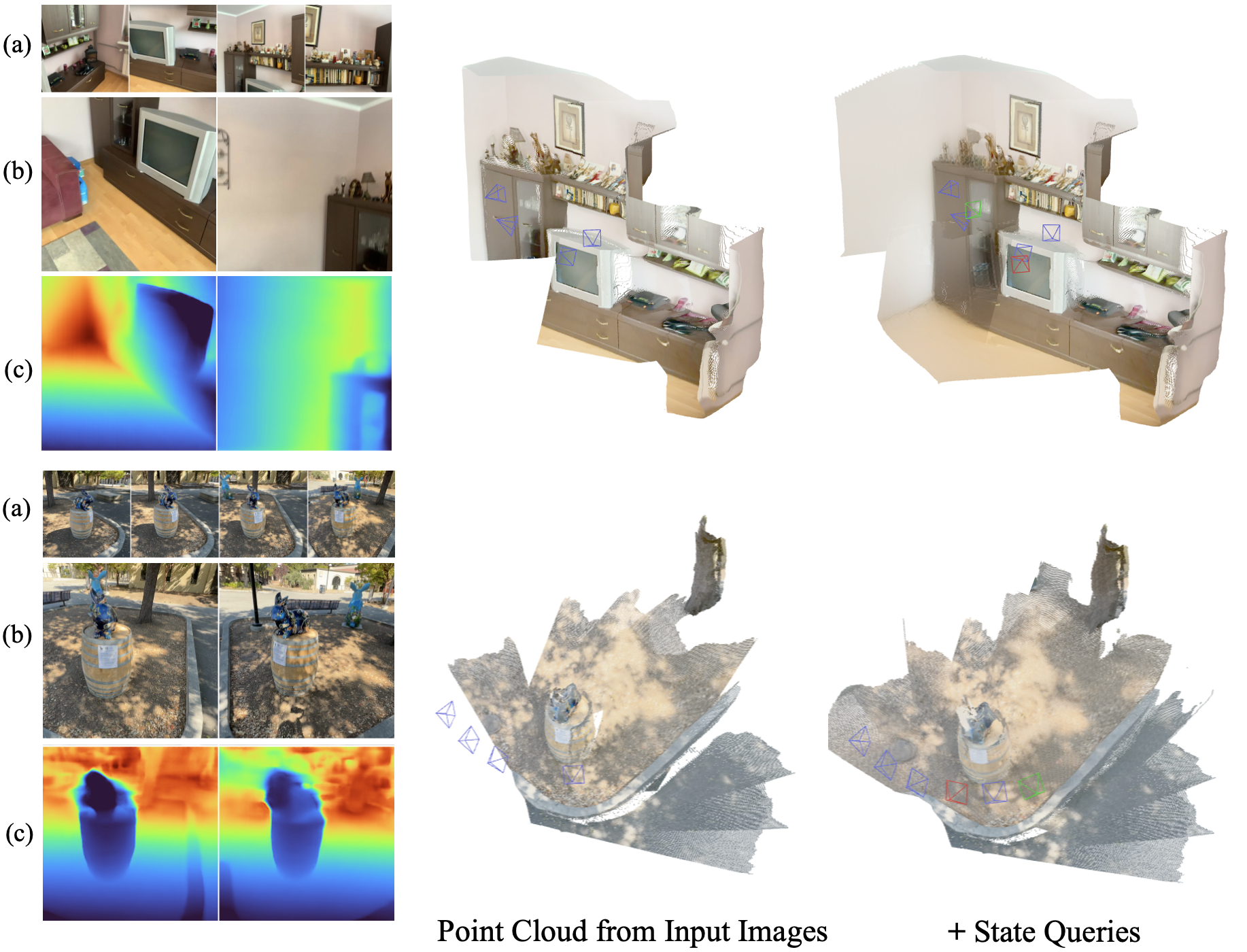}
    \vspace{-1em}
    \caption{\textbf{Querying Unseen Structure.} 
    Left: input images (a), GT images at query poses (b), and our predicted depth at those poses (c). Middle: point cloud reconstructed from input images only. Right: point cloud after querying (right column), where the queried point cloud is \textbf{merged} with the input image point cloud. This demonstrates our model's ability to infer common 3D structure (e.g.,\ walls, floors, and ground) in the unseen regions, thereby indicating an understanding of basic 3D scene priors.
    }
    \label{fig:query_unseen_structure}
\end{figure}

\subsection{Implicit Scene Representation} A unique capability of our model is that it compresses entire scene into a queryable hidden state via TTT layers. This state can be queried in real time ($\approx$100 FPS), independent of the input view number. We demonstrate this in two ways.

\medskip \noindent \textbf{Querying the Scene State Only.} In Figure~\ref{fig:query_seen_region} (Appendix), we show that the learned implicit scene state can be queried at novel-view camera poses to directly produce RGB and depth predictions. These predictions can be back-projected into 3D to form a colored point cloud. The resulting point cloud (right), obtained solely from state queries, closely matches the one reconstructed from the input images (middle), indicating that the learned scene state faithfully captures both the underlying geometry and appearance.

\medskip \noindent \textbf{Inferring Unseen Structure.} In Figure~\ref{fig:query_unseen_structure}, we demonstrate the model’s ability to infer plausible scene structure in unobserved regions. While its deterministic nature prevents it from hallucinating rich high-frequency details or entirely unseen objects (e.g., the sofa missing in the first example), it can still extrapolate common 3D structures such as walls, floors, and ground beyond the observed views, suggesting that the learned scene state encodes basic 3D scene priors.

\section{Conclusion}
We introduced {\methodname}, a stateful bidirectional architecture for feed-forward 3D reconstruction that scales in linear time. Across benchmarks, {\methodname} matches or surpasses the accuracy of state-of-the-art quadratic-time models, while being substantially faster.  Beyond efficiency, the learned scene state is queryable for real-time novel-view point-map prediction, and can be easily extended to streaming reconstruction application. Together, these results suggest a new path toward scalable, high-fidelity 3D perception on large image collections.

\newpage
\noindent \textbf{Acknowledgment}.  We would like to thank Shangzhan Zhang, Kyle Genova, Songyou Peng, and Zehao Yu for valuable discussions throughout the project. We thank Alfred Piccioni for help with setting up the training infrastructure, and Ben Poole for feedback on the manuscript. We also thank Yifan Wang and Jianyuan Wang for sharing baseline results and implementation details. Haian Jin was supported in part by a grant from the National Science Foundation (IIS-2211259) and by a Google PhD Fellowship.

{
    \small
    \bibliographystyle{ieeenat_fullname}
    \bibliography{main}
}


\input{supp_cleaned}

\end{document}

%% file: preamble.tex
\usepackage[export]{adjustbox}
\usepackage{xspace}
\usepackage{booktabs}  % For \toprule, \midrule, \cmidrule, \bottomrule
\usepackage{multirow}  % For \multirow
\usepackage{graphicx}  % For \resizebox
\usepackage{mathtools}
\usepackage{bm}
\usepackage{makecell}
\usepackage{placeins}
\usepackage{amsmath}   % For \uparrow (or \usepackage{amssymb})

\usepackage{fix-cm}

\usepackage[table]{xcolor}
\definecolor{tabfirst}{rgb}{1, 0.7, 0.7}
\definecolor{tabsecond}{rgb}{1, 0.85, 0.7}
\definecolor{tabthird}{rgb}{1, 1, 0.7}

\newcommand{\methodname}{ZipMap}

\usepackage{microtype}
\usepackage{siunitx}
\newcommand{\norm}[1]{\left\lVert#1\right\rVert}

\newcommand\lft{\mathopen{}\left}
\newcommand\rgt{\aftergroup\mathclose\aftergroup{\aftergroup}\right}

\newcommand{\vect}[1]{\bm{#1}}
\newcommand{\mat}[1]{{\mathbf{#1}}}

\newcommand{\point}{\vect{p}}
\newcommand{\wmat}{\mat{W}}
\newcommand{\wmatset}{\mathcal{W}}

\newcommand{\raymap}{T}
\newcommand{\camera}{\vect{c}}

\newcommand{\tablesize}{\footnotesize}

%% file: figures/tables.tex
\begin{table}[thb]
    \centering
    \vspace{-1em}
    \caption{\textbf{Camera Pose Estimation: RealEstate10K~\cite{zhou2018stereo} and Co3Dv2~\cite{reizenstein21co3d}.}
        We report pose AUC under angular error thresholds of 5/15/30 degrees. All methods have seen Co3Dv2 during training. CUT3R and TTT3R additionally use RealEstate10K for training, while the remaining methods do not.}
    % \vspace{-0.5em}
    
    \fontsize{7pt}{7pt}\selectfont
    \setlength{\tabcolsep}{0.5pt} 

    \begin{tabular}{l c c c c c c} 
        \toprule
        \multicolumn{1}{l}{\multirow{3}{*}{\textbf{Method}}} &
        \multicolumn{3}{c}{\textbf{RealEstate10K}} &
        \multicolumn{3}{c}{\textbf{Co3Dv2}} \\
        \cmidrule(lr){2-4} \cmidrule(lr){5-7}
        \multicolumn{1}{c}{} &
        AUC@5 $\uparrow$ & AUC@15 $\uparrow$ & AUC@30 $\uparrow$ &
        AUC@5 $\uparrow$ & AUC@15 $\uparrow$ & AUC@30 $\uparrow$ \\
        \midrule
        \multicolumn{7}{l}{\textit{$\mathcal{O}(N^2)$ Inference Speed}} \\      
        
        Fast3R~\cite{yang2025fast3r} & 22.36 & 46.71 & 61.68 & 31.05 & 59.63 & 73.43 \\
        FLARE~\cite{zhang2025flare}  & 38.47 & 67.02 & 80.01 & 23.84 & 57.78 & 73.99 \\
        VGGT~\cite{wang2025vggt}     & 38.71 & 66.46 & 78.89 & \cellcolor{tabfirst}67.84 & \cellcolor{tabfirst}83.95 & \cellcolor{tabfirst}89.99 \\
        $\pi^3$~\cite{wang2025pi3}   & \cellcolor{tabfirst}63.10 & \cellcolor{tabfirst}80.31 & \cellcolor{tabfirst}87.40 & \cellcolor{tabthird}57.12 & \cellcolor{tabthird}79.86 & \cellcolor{tabthird}87.93 \\
        \midrule
        \multicolumn{7}{l}{\textit{$\mathcal{O}(N)$ Inference Speed}} \\
        CUT3R~\cite{wang2025cut3r}   & \cellcolor{tabthird}46.92 & \cellcolor{tabthird}70.65 & \cellcolor{tabthird}81.68 & 24.88 & 56.28 & 71.72 \\
        TTT3R~\cite{chen2025ttt3r}   & 46.37 & 70.33 & 81.51 & 22.61 & 53.49 & 69.46 \\
        \textbf{Ours}                & \cellcolor{tabsecond}53.34 & \cellcolor{tabsecond}74.97 & \cellcolor{tabsecond}84.30 & \cellcolor{tabsecond}62.46 & \cellcolor{tabsecond}81.64 & \cellcolor{tabsecond}88.76 \\ 
        \bottomrule
    \end{tabular}
    \label{tab:camera_re10k_co3d}
\end{table}
\begin{table}[thb]
    \centering
    \caption{
        \textbf{Camera Pose Estimation: Sintel~\cite{bozic2021transformerfusion}, TUM-dynamics~\cite{Sturm2012ABF} and ScanNet~\cite{dai2017scannet}.} 
        We measure the distance
error of rotation/translation.
        All methods have seen ScanNet or ScanNet++~\cite{yeshwanth2023scannet++} in training. None has seen Sintel or TUM-dynamics.
    }
    \vspace{-0.5em}
    
    % 1. SET FONT SIZE
    % \scriptsize
    \fontsize{6pt}{7pt}\selectfont
    % 2. SET HORIZONTAL PADDING (Tight spacing)
    \setlength{\tabcolsep}{0.5pt}
    
    % 3. CLEAN COLUMN DEFINITION (1 'l' column + 9 'c' columns)
    \begin{tabular}{l c c c c c c c c c}
        \toprule
        {\multirow{3}{*}{\textbf{Method}}} &
        \multicolumn{3}{c}{\textbf{Sintel}} &
        \multicolumn{3}{c}{\textbf{TUM-dynamics}} &
        \multicolumn{3}{c}{\textbf{ScanNet}~(seen)} \\
        \cmidrule(lr){2-4} \cmidrule(lr){5-7} \cmidrule(lr){8-10}
        &
        ATE$\downarrow$ & RPE trans$\downarrow$ & RPE rot$\downarrow$ &
        ATE$\downarrow$ & RPE trans$\downarrow$ & RPE rot$\downarrow$ &
        ATE$\downarrow$ & RPE trans$\downarrow$ & RPE rot$\downarrow$ \\
        \midrule
        \multicolumn{7}{l}{\textit{$\mathcal{O}(N^2)$ Inference Speed}} \\      
        
        Fast3R~\cite{yang2025fast3r} & 0.371 & 0.298 & 13.750 & 0.090 & 0.101 & 1.425 & 0.155 & 0.123 & 3.491 \\

        FLARE~\cite{zhang2025flare}  & 0.207 & 0.090 & 3.015 & \cellcolor{tabthird}0.026 & 0.013 & 0.475 & 0.064 & 0.023 & 0.971 \\
        VGGT~\cite{wang2025vggt}     & \cellcolor{tabthird}0.172 & \cellcolor{tabsecond}0.061 & \cellcolor{tabsecond}0.471 & \cellcolor{tabfirst}0.012 & \cellcolor{tabsecond}0.010 & \cellcolor{tabsecond}0.309 & \cellcolor{tabthird}0.035 & \cellcolor{tabsecond}0.015 & \cellcolor{tabsecond}0.376 \\
        $\pi^3$~\cite{wang2025pi3}   & \cellcolor{tabfirst}0.073 & \cellcolor{tabfirst}0.038 & \cellcolor{tabfirst}0.288 & \cellcolor{tabsecond}0.014 & \cellcolor{tabfirst}0.009 & \cellcolor{tabfirst}0.307 & \cellcolor{tabfirst}0.030 & \cellcolor{tabfirst}0.013 & \cellcolor{tabfirst}0.345 \\
        \midrule
        \multicolumn{7}{l}{\textit{$\mathcal{O}(N)$ Inference Speed}} \\      
       CUT3R~\cite{wang2025cut3r}   & 0.216 & 0.071 & 0.622 & 0.042 & 0.013 & 0.395 & 0.096 & 0.022 & 0.578 \\
        TTT3R~\cite{chen2025ttt3r}   & 0.204 & 0.085 & 0.690 & 0.028 & \cellcolor{tabthird}0.012 & 0.361 & 0.065 & \cellcolor{tabthird}0.021 & 0.617 \\
        \textbf{Ours}                & \cellcolor{tabsecond}0.132 & \cellcolor{tabthird}0.066 & \cellcolor{tabsecond}0.438 & \cellcolor{tabfirst}0.012 & \cellcolor{tabsecond}0.010 & \cellcolor{tabthird}0.310 & \cellcolor{tabsecond}0.034 & \cellcolor{tabsecond}0.015 & \cellcolor{tabthird}0.385 \\
        \bottomrule
    \end{tabular}
    \label{tab:camera_sintel_tum_scannet}
\end{table}

\begin{table}[t]
    \centering
    \caption{
        \textbf{Point Map Estimation: 7-Scenes~\cite{Shotton2013SceneCR} and NRGBD~\cite{Azinovic_2022_CVPR}.} 
        Metrics are the same as in Table~\ref{tab:pointmap_dtu_eth3d}. Keyframes are selected every 200 (sparse) and 40 (dense) frames for 7-Scenes, and every 500 (sparse) and 100 (dense) frames for NRGBD. 
    }
    \vspace{-0.5em}
    
    % 1. Font Size & Padding
    \fontsize{6pt}{7pt}\selectfont
    \setlength{\tabcolsep}{1.7pt}

    \begin{tabular}{l c c c c c c c c c c c c}
        \toprule

        \multirow{3}{*}{\textbf{Method}} &
        \multicolumn{6}{c}{\textbf{7-Scenes}} &
        \multicolumn{6}{c}{\textbf{NRGBD}} \\
        \cmidrule(lr){2-7} \cmidrule(lr){8-13}
        
        % --- HEADER ROW 2 ---
        & \multicolumn{2}{c}{Acc. $\downarrow$}  &
        \multicolumn{2}{c}{Comp. $\downarrow$} &
        \multicolumn{2}{c}{NC. $\uparrow$}     & 
        \multicolumn{2}{c}{Acc. $\downarrow$}  &
        \multicolumn{2}{c}{Comp. $\downarrow$} &
        \multicolumn{2}{c}{NC. $\uparrow$}     \\
        \cmidrule(lr){2-3} \cmidrule(lr){4-5} \cmidrule(lr){6-7}
        \cmidrule(lr){8-9} \cmidrule(lr){10-11} \cmidrule(lr){12-13}
        
        % --- HEADER ROW 3 ---
        & Mean & Med. &
        Mean & Med. &
        Mean & Med. &
        Mean & Med. &
        Mean & Med. &
        Mean & Med. \\
        
        % ================= SPARSE SECTION =================
        \midrule
        \multicolumn{13}{l}{\textit{Sparse View}} \\
        Fast3R~\cite{yang2025fast3r} & 0.095 & 0.065 & 0.144 & 0.089 & 0.673 & 0.759 & 0.135 & 0.091 & 0.163 & 0.104 & 0.759 & 0.877 \\
        CUT3R~\cite{wang2025cut3r}   & \cellcolor{tabthird}0.080 & 0.055 & 0.102 & 0.066 & 0.711 & 0.811 & 0.098 & 0.038 & 0.075 & \cellcolor{tabthird}0.029 & 0.830 & 0.974 \\
        TTT3R~\cite{chen2025ttt3r}   & 0.098 & 0.062 & 0.159 & 0.107 & 0.681 & 0.768 & 0.101 & 0.039 & 0.076 & \cellcolor{tabthird}0.029 & 0.826 & 0.973 \\
        FLARE~\cite{zhang2025flare}  & 0.085 & 0.057 & 0.145 & 0.107 & 0.696 & 0.780 & 0.053 & \cellcolor{tabsecond}0.024 & \cellcolor{tabsecond}0.051 & \cellcolor{tabsecond}0.025 & 0.877 & \cellcolor{tabthird}0.988 \\
        VGGT~\cite{wang2025vggt}     & \cellcolor{tabfirst}0.044 & \cellcolor{tabfirst}0.024 & \cellcolor{tabfirst}0.056 & \cellcolor{tabfirst}0.033 & \cellcolor{tabthird}0.733 & \cellcolor{tabsecond}0.846 & \cellcolor{tabthird}0.049 & \cellcolor{tabthird}0.027 & 0.066 & 0.037 & \cellcolor{tabthird}0.882 & 0.979 \\
        $\pi^3$~\cite{wang2025pi3}   & \cellcolor{tabsecond}0.047 & \cellcolor{tabthird}0.029 & \cellcolor{tabthird}0.074 & \cellcolor{tabthird}0.049 & \cellcolor{tabfirst}0.741 & \cellcolor{tabthird}0.840 & \cellcolor{tabfirst}0.024 & \cellcolor{tabfirst}0.013 & \cellcolor{tabfirst}0.028 & \cellcolor{tabfirst}0.013 & \cellcolor{tabfirst}0.909 & \cellcolor{tabfirst}0.991 \\
        Ours                         & \cellcolor{tabfirst}0.044 & \cellcolor{tabsecond}0.026 & \cellcolor{tabsecond}0.065 & \cellcolor{tabsecond}0.037 & \cellcolor{tabsecond}0.740 & \cellcolor{tabfirst}0.853 & \cellcolor{tabsecond}0.046 & 0.028 & \cellcolor{tabthird}0.057 & 0.034 & \cellcolor{tabsecond}0.895 & \cellcolor{tabsecond}0.990 \\

        % ================= DENSE SECTION =================
        \midrule
        \multicolumn{13}{l}{\textit{Dense View}} \\
        Fast3R~\cite{yang2025fast3r} & 0.040 & 0.017 & 0.056 & 0.018 & 0.644 & 0.725 & 0.072 & 0.030 & 0.050 & 0.016 & 0.790 & 0.934 \\
        CUT3R~\cite{wang2025cut3r}   & 0.023 & \cellcolor{tabthird}0.010 & \cellcolor{tabthird}0.028 & \cellcolor{tabfirst}0.008 & 0.674 & 0.771 & 0.065 & 0.027 & 0.036 & 0.012 & 0.812 & 0.961 \\
        TTT3R~\cite{chen2025ttt3r}   & 0.035 & 0.016 & 0.032 & \cellcolor{tabsecond}0.010 & 0.666 & 0.760 & 0.074 & 0.033 & 0.037 & 0.014 & 0.803 & 0.957 \\
        FLARE~\cite{zhang2025flare}  & \cellcolor{tabthird}0.019 & \cellcolor{tabfirst}0.007 & \cellcolor{tabsecond}0.026 & 0.013 & \cellcolor{tabsecond}0.684 & \cellcolor{tabsecond}0.785 & 0.023 & 0.011 & 0.018 & \cellcolor{tabthird}0.008 & \cellcolor{tabfirst}0.882 & \cellcolor{tabfirst}0.986 \\
        VGGT~\cite{wang2025vggt}     & 0.022 & \cellcolor{tabsecond}0.008 & \cellcolor{tabsecond}0.026 & 0.012 & 0.667 & 0.760 & \cellcolor{tabsecond}0.015 & \cellcolor{tabsecond}0.008 & \cellcolor{tabsecond}0.015 & \cellcolor{tabfirst}0.006 & \cellcolor{tabthird}0.871 & \cellcolor{tabthird}0.982 \\
        $\pi^3$~\cite{wang2025pi3}   & \cellcolor{tabfirst}0.016 & \cellcolor{tabfirst}0.007 & \cellcolor{tabfirst}0.022 & \cellcolor{tabthird}0.011 & \cellcolor{tabfirst}0.689 & \cellcolor{tabfirst}0.792 & \cellcolor{tabfirst}0.013 & \cellcolor{tabfirst}0.007 & \cellcolor{tabfirst}0.014 & \cellcolor{tabfirst}0.006 & \cellcolor{tabsecond}0.874 & 0.981 \\
        Ours                         & \cellcolor{tabsecond}0.018 & \cellcolor{tabsecond}0.008 & 0.030 & 0.012 & \cellcolor{tabthird}0.680 & \cellcolor{tabthird}0.780 & \cellcolor{tabthird}0.016 & \cellcolor{tabthird}0.009 & \cellcolor{tabthird}0.017 & \cellcolor{tabsecond}0.007 & 0.870 & \cellcolor{tabsecond}0.983 \\
        \bottomrule
    \end{tabular}
    \label{tab:pointmap_7s_nrgbd}
\end{table}

\begin{table}[t]
    \centering
    \caption{
        \textbf{Point Map Estimation: DTU~\cite{Jensen2014LargeSM} and ETH3D~\cite{schoeps2017cvpr}.} 
        Following $\pi^3$~\cite{wang2025pi3}, we report the mean and median of accuracy (Acc.), completeness (Comp.), and normal-consistency (NC.) for keyframes selected every 5 images.
    }
    \vspace{-0.5em}
    
    % 1. Font Size
    \fontsize{6pt}{7pt}\selectfont
    
    % 2. Tight Padding
    \setlength{\tabcolsep}{1.7pt}
    
    % 3. Clean Column Definition (1 Method + 12 Metrics = 13 Columns)
    \begin{tabular}{l c c c c c c c c c c c c}
        \toprule
        
        % --- HEADER ROW 1 ---
        \multirow{3}{*}{\textbf{Method}} &
        \multicolumn{6}{c}{\textbf{DTU}} &
        \multicolumn{6}{c}{\textbf{ETH3D}} \\
        \cmidrule(lr){2-7} \cmidrule(lr){8-13}
        
        % --- HEADER ROW 2 ---
        & \multicolumn{2}{c}{Acc. $\downarrow$}  &
        \multicolumn{2}{c}{Comp. $\downarrow$} &
        \multicolumn{2}{c}{N.C. $\uparrow$}    & 
        \multicolumn{2}{c}{Acc. $\downarrow$}  &
        \multicolumn{2}{c}{Comp. $\downarrow$} &
        \multicolumn{2}{c}{N.C. $\uparrow$}    \\
        \cmidrule(lr){2-3} \cmidrule(lr){4-5} \cmidrule(lr){6-7}
        \cmidrule(lr){8-9} \cmidrule(lr){10-11} \cmidrule(lr){12-13}
        
        % --- HEADER ROW 3 ---
        & Mean & Med. &
        Mean & Med. &
        Mean & Med. &
        Mean & Med. &
        Mean & Med. &
        Mean & Med. \\
        \midrule
        \multicolumn{7}{l}{\textit{$\mathcal{O}(N^2)$ Inference Speed}} \\      
        Fast3R~\cite{yang2025fast3r} & 3.340 & 1.919 & 2.929 & 1.125 & \cellcolor{tabthird}0.671 & \cellcolor{tabthird}0.755 & 0.832 & 0.691 & 0.978 & 0.683 & 0.667 & 0.766 \\
        FLARE~\cite{zhang2025flare}  & 2.541 & 1.468 & 3.174 & 1.420 & \cellcolor{tabfirst}0.684 & \cellcolor{tabfirst}0.774 & 0.464 & 0.338 & 0.664 & 0.395 & 0.744 & 0.864 \\
        VGGT~\cite{wang2025vggt}     & \cellcolor{tabthird}1.308 & \cellcolor{tabthird}0.761 & \cellcolor{tabthird}1.929 & \cellcolor{tabthird}1.015 & 0.665 & 0.751 & \cellcolor{tabthird}0.270 & \cellcolor{tabthird}0.174 & \cellcolor{tabthird}0.304 & \cellcolor{tabthird}0.180 & \cellcolor{tabthird}0.841 & \cellcolor{tabthird}0.942 \\
        $\pi^3$~\cite{wang2025pi3}   & \cellcolor{tabfirst}1.151 & \cellcolor{tabfirst}0.622 & \cellcolor{tabsecond}1.793 & \cellcolor{tabfirst}0.629 & 0.668 & 0.754 & \cellcolor{tabfirst}0.188 & \cellcolor{tabfirst}0.126 & \cellcolor{tabfirst}0.211 & \cellcolor{tabfirst}0.129 & \cellcolor{tabfirst}0.872 & \cellcolor{tabfirst}0.967 \\
        \midrule
        \multicolumn{7}{l}{\textit{$\mathcal{O}(N)$ Inference Speed}} \\      

        CUT3R~\cite{wang2025cut3r}   & 5.045 & 2.954 & 6.437 & 4.146 & 0.666 & 0.742 & 0.593 & 0.461 & 0.747 & 0.590 & 0.754 & 0.863 \\
        TTT3R~\cite{chen2025ttt3r}   & 5.337 & 3.261 & 6.593 & 4.236 & 0.666 & 0.743 & 0.763 & 0.633 & 0.881 & 0.617 & 0.739 & 0.840 \\      
        \textbf{Ours}                & \cellcolor{tabsecond}1.228 & \cellcolor{tabsecond}0.671 & \cellcolor{tabfirst}1.649 & \cellcolor{tabsecond}0.663 & \cellcolor{tabsecond}0.675 & \cellcolor{tabsecond}0.764 & \cellcolor{tabsecond}0.254 & \cellcolor{tabsecond}0.171 & \cellcolor{tabsecond}0.249 & \cellcolor{tabsecond}0.159 & \cellcolor{tabsecond}0.865 & \cellcolor{tabsecond}0.965 \\
        \bottomrule
    \end{tabular}
    \label{tab:pointmap_dtu_eth3d}
\end{table}

\begin{table}[thb]
    \centering
    \caption{
        \textbf{Video Depth Estimation: Sintel~\cite{bozic2021transformerfusion}, Bonn~\cite{Palazzolo2019ReFusion3R} and KITTI~\cite{Geiger2013IJRR}.} We report results under \textbf{scale-only} alignment; results using joint scale-and-shift alignment are provided in the Appendix.
    }
    \vspace{-1em}
    
    % 1. Font Size & Padding
    \fontsize{6.3pt}{7pt}\selectfont
    \setlength{\tabcolsep}{1pt} % Tight padding
    
    % 2. Column Definition: Method(l) + Params(c) + 6 Metrics(c) = 8 Columns
    \begin{tabular}{l c c c c c c c}
        \toprule
        
        \multicolumn{2}{c}{} &
        \multicolumn{2}{c}{\textbf{Sintel}} &
        \multicolumn{2}{c}{\textbf{Bonn}} &
        \multicolumn{2}{c}{\textbf{KITTI}} \\
        
        % cmidrules cover the metric pairs (cols 3-4, 5-6, 7-8)
        \cmidrule(lr){3-4} \cmidrule(lr){5-6} \cmidrule(lr){7-8}
        
        % --- HEADER ROW 2 ---
        \textbf{Method} &
        \textbf{Params} &
        AbsRel$\downarrow$ & $\delta<1.25$$\uparrow$ &
        AbsRel$\downarrow$ & $\delta<1.25$$\uparrow$ &
        AbsRel$\downarrow$ & $\delta<1.25$$\uparrow$ \\
        
        \midrule
        \multicolumn{8}{l}{\textit{$\mathcal{O}(N^2)$ Inference Speed}} \\
        % \midrule
        
        Fast3R~\cite{yang2025fast3r}     & 648M & 0.638 & 0.422 & 0.194 & 0.772 & 0.138 & 0.834 \\
        FLARE~\cite{zhang2025flare}      & 1.40B & 0.729 & 0.336 & 0.152 & 0.790 & 0.356 & 0.570 \\
        VGGT~\cite{wang2025vggt}         & 1.26B & \cellcolor{tabthird}0.298 & \cellcolor{tabthird}0.643 & \cellcolor{tabsecond}0.055 & \cellcolor{tabthird}0.971 & 
        \cellcolor{tabthird}0.073 & 
        \cellcolor{tabthird}0.965 \\
        $\pi^3$~\cite{wang2025pi3}       & 959M & \cellcolor{tabfirst}0.228 & \cellcolor{tabsecond}0.672 & \cellcolor{tabfirst}0.051 & \cellcolor{tabfirst}0.975 & \cellcolor{tabfirst}0.038 & \cellcolor{tabfirst}0.986 \\

        \midrule
        \multicolumn{8}{l}{\textit{$\mathcal{O}(N)$ Inference Speed}} \\ 
        CUT3R~\cite{wang2025cut3r}       & 793M & 0.432 & 0.510 & 0.072 & 0.951 & 0.152 & 0.805 \\
        TTT3R~\cite{chen2025ttt3r}       & 793M & 0.426 & 0.522 & 0.061 & 0.970 & 0.149 & 0.812 \\
        Ours                             & 1.40B & \cellcolor{tabsecond}0.248 & \cellcolor{tabfirst}0.695 & \cellcolor{tabthird}0.059 & \cellcolor{tabsecond}0.973 & \cellcolor{tabsecond}0.057 & \cellcolor{tabsecond}0.974 \\
        \bottomrule
    \end{tabular}
    \label{tab:videodepth_scale}
\end{table}

%% file: supp_cleaned.tex
% \input{preamble}
% Define Colors
\definecolor{citecolor}{HTML}{0071bc} 
\definecolor{tblue}{RGB}{80,80,245}
\definecolor{tred}{RGB}{250,100,100}
\definecolor{tgreen}{RGB}{32,178,170}
\definecolor{tgray}{RGB}{169,169,169}
\definecolor{tbg}{RGB}{230,245,230}
\definecolor{tbb}{RGB}{135,206,250}
\definecolor{blue2}{RGB}{0,139,139}
\definecolor{red2}{RGB}{255,127,80}
\definecolor{NiceBlue}{RGB}{80,80,245} % Added definition for tablefirst
\definecolor{lightyellow}{rgb}{1,1, 0.8}
\definecolor{yellow}{rgb}{1,0.97, 0.65}
\definecolor{orange}{rgb}{1, 0.85, 0.7}
\definecolor{tablered}{rgb}{1, 0.7, 0.7}
\definecolor{tablegreen}{rgb}{0.80, 1, 0.80}
\definecolor{tablered2}{rgb}{0.8, 0.8, 1.0}

% --- Hyperref (Must be loaded last) ---

% --- Custom Commands ---
\newcommand{\ve}[1]{\mathbf{#1}} % for displaying a vector
\newcommand{\ma}[1]{\mathrm{#1}} % for displaying a matrix

\newcommand{\rbg}[1]{{\color{blue}[rbg: #1]}}
\newcolumntype{x}[1]{>{\centering\arraybackslash}p{#1pt}}

\newcommand{\bd}[1]{\textbf{#1}}
\newcommand{\app}{\raise.17ex\hbox{$\scriptstyle\sim$}}
\newcommand{\ncdot}{{\mkern 0mu\cdot\mkern 0mu}}
\def\x{$\times$}
\newcommand{\dt}[1]{\fontsize{8pt}{.1em}\selectfont \emph{#1}}

% Table styling
\newlength\savewidth
\newcommand\shline{\noalign{\global\savewidth\arrayrulewidth
  \global\arrayrulewidth 1pt}\hline\noalign{\global\arrayrulewidth\savewidth}}

\newcommand{\customfootnotetext}[2]{{% Group to localize change to footnote
  \renewcommand{\thefootnote}{#1}% Update footnote counter representation
  \footnotetext[0]{#2}}}% Print footnote text

\newcommand{\padcell}{\cellcolor{citecolor!10}}
\newcommand{\padcellred}{\cellcolor{red!10}}

\newcommand{\green}[1]{{\color{tgreen}#1}}

% Checkmarks
\newcommand{\cmark}{\color{tgreen}\ding{51}}%
\newcommand{\xmark}{\color{red2}\ding{55}}%

\newcommand{\tablefirst}{\cellcolor{NiceBlue!25}}

\newcommand{\modelname}{FRM}

%%%%%%%%% PAPER ID  - PLEASE UPDATE
% \def\paperID{210} 
\def\confName{CVPR}
\def\confYear{2026}

% \begin{document}

\maketitlesupplementary

\newpage
\section*{Outline}
In this Supplementary Material, we provide the following:

\begin{itemize}
    \item \textbf{Appendix~\ref{appendix:eval_details}: Evaluation Details.} Comprehensive details on baseline evaluation, runtime evaluation, long-sequence evaluation protocols.
    \item \textbf{Appendix~\ref{appendix:training_details}: Training Details.} Full descriptions of the training datasets, the complete training loss function, and additional implementation details for finetuning the model for scene state query and streaming reconstruction.
    \item \textbf{Appendix~\ref{appendix:state_query}: More Results for the Implicit Scene State.} Visualizations demonstrating the ability to query the learned implicit scene state (Figure~\ref{fig:query_seen_region}).
    \item \textbf{Appendix~\ref{appendix:more_eval_results}: More Evaluation Results.} Additional quantitative and qualitative results, including monocular depth estimation benchmarks (Table~\ref{tab:monodepth}), general qualitative comparisons (Figure~\ref{fig:qualitative_comparison}), effects of removing the reference view (Figure~\ref{fig:ablation_ref_longseq}, Tables~\ref{tab:camera_sintel_tum_scannet_ref_view}, \ref{tab:videodepth_ref_view}, \ref{tab:pointmap_dtu_eth3d_ref_view}), and more long-sequence evaluation results (Figure~\ref{fig:longseq_eval_videodepth}, Figure~\ref{fig:longseq_eval_7scenes}).
    \item \textbf{Appendix~\ref{appendix:limitations}: Limitations.} A discussion of the current limitations of our proposed method.
\end{itemize}

\appendix

\section{Evaluation Details}
\label{appendix:eval_details}
\subsection{Baseline Evaluation Details}
To produce the results in Section 4 of the main paper, we evaluated CUT3R~\cite{wang2025cut3r}, TTT3R~\cite{chen2025ttt3r}, VGGT~\cite{wang2025vggt}, $\pi^3$~\cite{wang2025pi3}, and our method directly. For all other baselines, we use the results reported in the $\pi^3$ paper. We resize input images according to patch size: for CUT3R and TTT3R (patch size $16$), we set the image width to $512$, while for VGGT, $\pi^3$, and our method (patch size $14$), we set the image width to $518$.

\subsection{Runtime Evaluation Details}
\label{appendix:runtime_eval}
\begin{table*}[h]
    \centering
    \caption{\textbf{Runtime evaluation.} Inference time (in seconds) as a function of the number of input images $N$. VGGT and $\pi^3$ scale quadratically with $N$, resulting in slow speeds when $N$ is large. CUT3R, TTT3R, and our method scale linearly with $N$, with ours being the fastest for dense input frames.}
    \label{tab:speed_test}
    \resizebox{\linewidth}{!}{
    \begin{tabular}{lccrrrrrrrrrr} 
        \toprule
        \textbf{Model} & \textbf{Complexity} & \textbf{Params} & \multicolumn{10}{c}{\textbf{Time to Reconstruct $N$ Frames (seconds)}} \\
        \cmidrule(lr){4-13}
        & & & $N =$ 5 & 10 & 25 & 50 & 100 & 200 & 300 & 400 & 500 & 750 \\
        \midrule
        VGGT~\cite{wang2025vggt} & $\mathcal{O}(N^2)$ & 1.26B & 0.102 & 0.194 & 0.569 & 1.524 & 4.689 & 16.040 & 34.022 & 58.842 & 90.389 & 200.364 \\
        $\pi^3$~\cite{wang2025pi3} & $\mathcal{O}(N^2)$ & 959M & 0.087 & 0.157 & 0.450 & 1.186 & 3.604 & 12.190 & 25.765 & 44.464 & 68.255 & 151.159 \\
        CUT3R~\cite{wang2025cut3r} & $\mathcal{O}(N)$ & 793M & 0.206 & 0.413 & 1.018 & 2.056 & 4.088 & 8.222 & 12.430 & 16.618 & 21.025 & 31.246 \\
        TTT3R~\cite{chen2025ttt3r} & $\mathcal{O}(N)$ & 793M & 0.206 & 0.411 & 1.033 & 2.036 & 4.128 & 8.267 & 12.435 & 16.511 & 20.767 & 31.197 \\
        \textbf{Ours} & $\mathcal{O}(N)$ & 1.40B & 0.125 & 0.183 & 0.383 & 0.712 & 1.362 & 2.681 & 4.017 & 5.348 & 6.671 & 9.999 \\
        \bottomrule
    \end{tabular}
    }
\end{table*}
\begin{figure*}[thb]
    \centering
    \includegraphics[width=\linewidth]{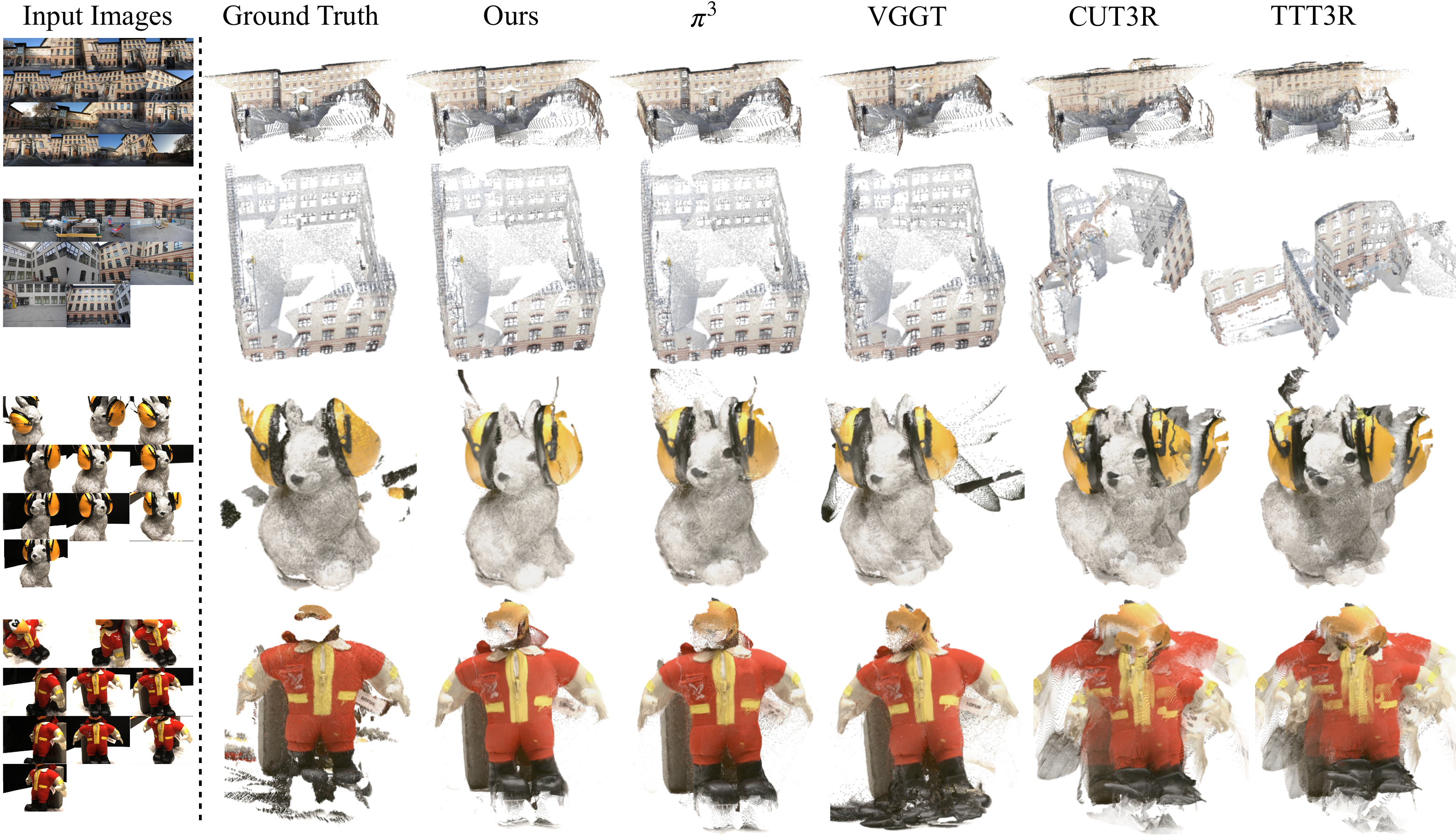}
    \caption{\textbf{Qualitative comparison.} Point cloud reconstructions of scenes from the ETH3D and DTU datasets.}
    \label{fig:qualitative_comparison}
    % \vspace{1em}
\end{figure*}
To produce the runtime analysis shown in Figure~1 of the main paper, we evaluate all methods on a single H100 SXM5 GPU using PyTorch 2.7.1 and CUDA 12.8. All implementations use the pytorch \texttt{scaled\_dot\_product\_attention} function for softmax attention, which is a cuDNN implementation of FlashAttention-2~\cite{dao2023flashattention2}. Input resolutions follow the same aspect ratio as the ScanNet-v2~\cite{dai2017scannet} dataset used to evaluate the pose estimation error in Figure~1: $392 \times 518$ for VGGT, $\pi^3$, and our method (patch size $14$), and $384 \times 512$ for CUT3R and TTT3R (patch size $16$). The original VGGT implementation ran out of GPU memory on long input sequences (e.g.,\ $750$ frames) because it caches features from all layers, even though the DPT heads only use four of them. To enable long-sequence evaluation, we optimized VGGT's implementation to store only the features required by the DPT heads, which eliminates these out-of-memory issues without affecting accuracy or runtime. We evaluate sequences up to $750$ frames, as this is already close to the memory limit (80GB) of both our method and the baseline~\cite{wang2025vggt}. The reported runtime is averaged over 10 iterations, after 2 warm-up iterations. See Table~\ref{tab:speed_test} for detailed runtimes.

As we see in Table~\ref{tab:speed_test}, when the input views are very sparse (e.g.,\ $5$ frames), all methods are able to complete reconstruction quickly. Our method is slightly slower than the quadratic methods (VGGT and $\pi^3$), 
likely because (i) our method implements the test-time training block using standard PyTorch code, whereas the quadratic baselines rely on highly optimized fused FlashAttention kernels, and (ii) our method applies Newton--Schulz orthonormalization during the forward pass (Equation 4 in the main paper), which incurs a constant additional cost. As the number of input frames increases, our method exhibits a clear speed advantage. At $750$ frames, our method finishes reconstruction in under $10$ seconds (75 FPS) , which is more than $20\times$ faster than VGGT and $15\times$ faster than $\pi^3$.

When querying the implicit scene state, we only do the apply operation to the TTT blocks without performing any update step. As a result, querying is faster than reconstruction, reaching about 100 FPS in our experiments.

\subsection{Long-Sequence Evaluation Details}
We follow the evaluation protocol in $\pi^3$~\cite{wang2025pi3} and take the first $N$ frames of each test sequence of ScanNet-v2~\cite{dai2017scannet} dataset for evaluating camera pose estimation and video depth estimation. We evaluated up to $N=750$ frames on ScanNet-v2.
We also follow the evaluation protocol in $\pi^3$ for evaluating 3D point estimation on 7-Scenes~\cite{Shotton2013SceneCR} dataset, by either taking the first $N$ frames or uniformly subsampling $N$ frames. Due to the slow ICP alignment process when computing the chamfer distance, we only evaluated up to $N=300$ frames under time constraint.
For the evaluation of camera pose estimation on DL3DV~\cite{ling2024dl3dv} ($55$ scenes in test split), we additionally exclude $5\%$ of scenes with the largest errors when calculating the metrics for each method to mitigate the impact of outliers. We only evaluated up to $N=300$ frames since most of DL3DV scenes have no more than $400$ frames.

\section{Training Details}
\label{appendix:training_details}
\subsection{Full Training Datasets}
\label{appendix:training_datasets}
We train our model on a diverse collection of 29 publicly available datasets.
We use 23 static scene datasets, including Aria Synthetic Environments~\cite{pan2023aria}, ARKitScenes~\cite{baruch2021arkitscenes}, BlendedMVS~\cite{yao2020blendedmvs}, Co3dv2~\cite{reizenstein21co3d}, DL3DV~\cite{ling2024dl3dv}, GTA-SfM~\cite{wang2020gta_sfm}, Hypersim~\cite{roberts2021hypersim}, MapFree~\cite{arnold2022map}, Matrixcity~\cite{li2023matrixcity}, Matterport3D~\cite{Matterport3D}, MegaDepth~\cite{megadepth}, MidAir~\cite{Fonder2019MidAir}, MVS-Synth~\cite{mvssynth}, OmniObject3D~\cite{wu2023omniobject3d}, ScanNet~\cite{dai2017scannet}, ScanNet++~\cite{yeshwanth2023scannet++}, ScenenetRGBD~\cite{mccormac2017scenenet}, TartanAir~\cite{tartanair2020iros}, TartanGround
~\cite{patel2025tartanground}, Unreal4k~\cite{zhang2018unrealstereo}, Virtual KITTI~\cite{cabon2020virtualkitti2}, Waymo~\cite{sun2020waymo}, WildRGBD.
We also use 6 dynamic scene datasets, including BEDLAM~\cite{BEDLAM_Black_CVPR_2023}, Dynamic Replica~\cite{karaev2023dynamicstereo}, Kubric~\cite{greff2022kubric}, OmniWorld~\cite{zhou2025omniworld}, PointOdyssey~\cite{zheng2023pointodyssey}, and Spring~\cite{mehl2023spring}.

\subsection{Full Training Loss}
\label{appendix:full_training_loss}
In addition to the training loss described in the main paper, we also use a normal loss $\mathcal{L}_{\text{point-normal}}$ to supervise local point map prediction, and a gradient loss $\mathcal{L}_{\text{depth-grad}}$ to regularize predicted depths to be locally smooth:
\begin{gather}
\mathcal{L}_{\text{point-normal}} = \operatorname*{mean}_{i, j}\lft( \arccos\lft(\mathbf{n}_{i,j} \cdot \mathbf{n}^*_{i,j}\rgt)\rgt)
\label{eq:point_normal_loss}\,, \\
\mathcal{L}_{\text{depth-grad}} = \operatorname*{mean}_i \lft( \norm{\Sigma_i \circ \lft(\nabla \lft(\hat{s} D_i\rgt) - \nabla D^*_i \rgt)}_1 \rgt)\,,
\label{eq:depth_grad_loss}
\end{gather}
where the normal $\mathbf{n}_{i,j}$ of pixel $j$ from view $i$ is obtained by computing the cross product of its adjacent edges on the predicted local point map, and $\mathbf{n}^*_{i,j}$ is computed from the ground truth point map with the same procedure.

\subsection{More Implementation Details}
\label{appendix:more_imple_details}
\medskip \noindent \textbf{Training Implementation.} 
Our model is trained with Fully Sharded Data Parallel (FSDP), and we apply \texttt{torch.compile} to the test-time training block to accelerate training. Following VGGT~\cite{wang2025vggt}, we randomly apply color jitter, Gaussian blur, and grayscale to the input frames as data augmentation. For training stability, we normalize the ground-truth cameras, depths, and local points using the global point cloud scale.  During training, input images are resized to a width of $518$ pixels with a random aspect ratio sampled from $[0.33, 1.0]$. For each scene, we randomly sample $2$–$48$ frames and cap the number of images per GPU at $48$.

\medskip \noindent \textbf{Finetune the Model for Scene State Query.} 
We fine-tune the trained model to enable scene state queries. We use the first input frame as the reference view and express the target query camera pose in the coordinate system of this reference view. Our camera prediction is scale-invariant, but we need to fix the scale of the target camera to improve training stability. Therefore, the scale of the target camera is different from the predicted cameras. Specifically, we determine the target camera translation scale from the maximum distance of all input camera centers to the origin (the camera center of the reference view).

We fine-tune the model for $100\text{K}$ iterations. During fine-tuning, we keep all other training losses unchanged and additionally include the query losses described in the main paper. Since the RGB loss requires photometrically consistent inputs, we disable color-based data augmentation (color jitter, Gaussian blur, and grayscale) on the input frames. In addition, we exclude dynamic datasets and static datasets with inconsistent image collections, such as MegaDepth~\cite{megadepth}.
We observed that the LPIPS loss introduces substantial extra GPU memory overhead. Therefore, during fine-tuning we reduce the maximum number of images per GPU from $48$ to $44$. For each scene, we randomly sample $4$–$44$ frames and use half of them as input frames and the other half as target frames. Consequently, each training example uses $2$–$22$ input views, and the number of target views matches the number of input views.

\medskip \noindent \textbf{Finetune the Model for Streaming Reconstruction.} 
To enable streaming reconstruction, we replace the transformer-based camera head with a lightweight two-layer MLP, and finetune the Stage-3 checkpoint (trained without an explicit reference view) on 32 H100 GPUs. We train the model on all datasets used before.
We first train it for 60k steps with a learning rate of $1\mathrm{e}{-5}$, using 36 images per GPU (12 images per scene), and then continue for another 30k steps at the same learning rate with longer context (24 images per scene), using 48 images per GPU.
\textit{We observe a notable gain when increasing the training context from 12 to 24 views.}
Due to time constraints we stop at 24 views; however, since our streaming baselines are trained with longer contexts (up to 64 views), we reasonably expect further scaling the context length toward 64 views to yield an even larger advantage.

\begin{figure*}[thb]
    \centering
    \includegraphics[width=0.90\linewidth]{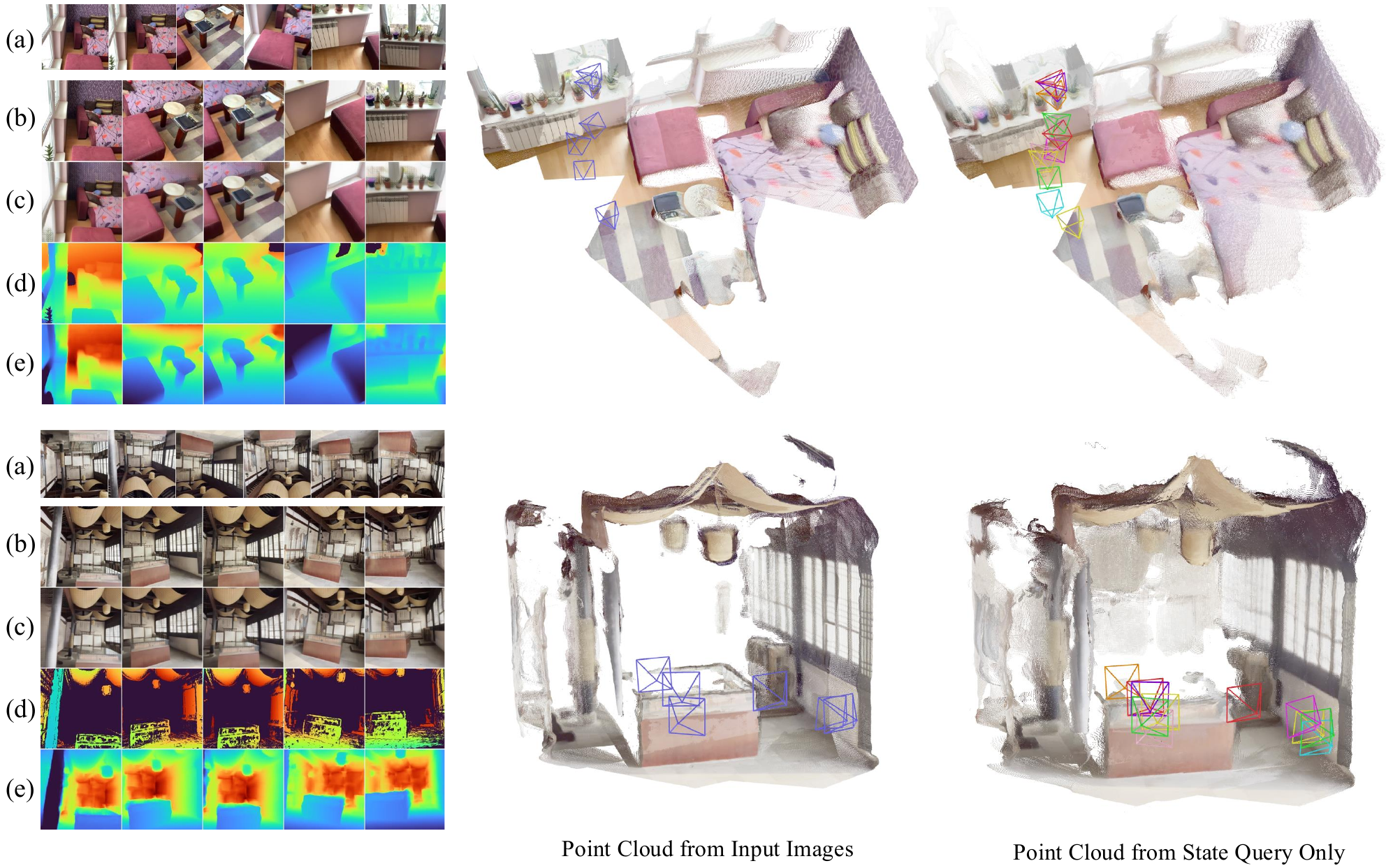}
    \vspace{-8pt}
    \caption{\textbf{Querying the Scene State.} The left panels show: input images (a), GT RGB at query poses (b), our RGB predictions (c), GT depth (d), and predicted depth (e). The middle panels visualize the 3D point clouds reconstructed from the input images. The right panels show point clouds attained \textbf{solely by querying the scene state}. The close visual match between these two point clouds indicates that the learned scene state faithfully captures the geometry and appearance of the input scene.}
    \label{fig:query_seen_region}
\end{figure*}

\section{More Results for the Implicit Scene State}
\label{appendix:state_query}
In Figure~\ref{fig:query_unseen_structure} of the main paper, we demonstrate our model’s ability to infer scene structure in unseen regions.

In Figure~\ref{fig:query_seen_region}, we further show that the reconstructed implicit scene state can be directly queried at novel view camera poses to obtain RGB and depth predictions. These predictions can then be back-projected into 3D to form colored point cloud. Notably, the point cloud attained solely from state queries closely resembles the geometry and appearance of the point cloud reconstructed from the input images, indicating that the learned scene state faithfully captures both the geometry and appearance of the underlying scene.

\section{More Evaluation Results}
\label{appendix:more_eval_results}

\begin{table}[htb]
    \centering
    \caption{
        \textbf{Monocular Depth Estimation.} We evalute the frame-independent monocular depth estimation on the Sintel~\cite{bozic2021transformerfusion}, Bonn~\cite{Palazzolo2019ReFusion3R}, KITTI~\cite{Geiger2013IJRR} and NYU-v2~\cite{silberman2012indoor} datasets.
        \label{tab:monodepth}
        } 
    \vspace{-0.5em}
    % \tablestyle{1pt}{1.05}
    \resizebox{\linewidth}!{
    % \tablesize
    \begin{tabular}{lcccccccc}

         &
        \multicolumn{2}{c}{\textbf{Sintel}} &
        \multicolumn{2}{c}{\textbf{Bonn}} &
        \multicolumn{2}{c}{\textbf{KITTI}} &
        \multicolumn{2}{c}{\textbf{NYU-v2}} \\
        \cmidrule(r){2-3} \cmidrule(r){4-5} \cmidrule(r){6-7} \cmidrule(r){8-9}
        \textbf{Method} &
        AbsRel$\downarrow$ & $\delta<1.25$$\uparrow$ &
        AbsRel$\downarrow$ & $\delta<1.25$$\uparrow$ &
        AbsRel$\downarrow$ & $\delta<1.25$$\uparrow$ &
        AbsRel$\downarrow$ & $\delta<1.25$$\uparrow$ \\
        \midrule%
MASt3R~\cite{leroy2024grounding} &                      0.413 &                      0.569 &                      0.123 &                      0.833 &                      0.077 &                      0.948 &                      0.110 &                      0.865 \\
MonST3R~\cite{zhang2024monst3r}  &                      0.402 &                      0.525 &                      0.069 &                      0.954 &                      0.098 &                      0.895 &                      0.094 &                      0.887 \\
Fast3R~\cite{yang2025fast3r}     &                      0.544 &                      0.509 &                      0.169 &                      0.796 &                      0.120 &                      0.861 &                      0.093 &                      0.898 \\
CUT3R~\cite{wang2025cut3r}       &                      0.418 &                      0.520 &                      0.058 &                      0.967 &                      0.097 &                      0.914 &                      0.081 &                      0.914 \\
FLARE~\cite{zhang2025flare}      &                      0.606 &                      0.402 &                      0.130 &                      0.836 &                      0.312 &                      0.513 &                      0.089 &                      0.898 \\

MoGe v1~\cite{wang2025moge}           &  \cellcolor{tabsecond}0.273 &   \cellcolor{tabfirst}0.695 &   \cellcolor{tabfirst}0.050 &   \cellcolor{tabfirst}0.976 &  0.054 &  
                   0.977 &    \cellcolor{tabthird}0.055 &                       0.952 \\
MoGe v2~\cite{wang2025mogev2}         &                       0.277 &  0.687 &                       0.063 &   0.973 &    \cellcolor{tabfirst}0.049 &   \cellcolor{tabfirst}0.979 &                       0.060 &                       0.940 \\
VGGT~\cite{wang2025vggt}         &                      0.329 &                      0.600 & \cellcolor{tabsecond}0.051 & \cellcolor{tabsecond}0.974 &                      0.089 &                      0.939 &  \cellcolor{tabthird}0.055 &  \cellcolor{tabthird}0.953 \\
$\pi^3$~\cite{wang2025pi3}       &  \cellcolor{tabthird}0.276 &   \cellcolor{tabthird}0.622 &  \cellcolor{tabthird}0.052 &                      0.971 &  \cellcolor{tabsecond}0.059 &  \cellcolor{tabsecond}0.972 & \cellcolor{tabsecond}0.054 & \cellcolor{tabsecond}0.956 \\
\textbf{Ours}                    &  \cellcolor{tabfirst}0.268 &  \cellcolor{tabsecond}0.666 &                      0.056 &  \cellcolor{tabthird}0.973 &  \cellcolor{tabthird}0.063 &  \cellcolor{tabthird}0.960 &  \cellcolor{tabfirst}0.052 &  \cellcolor{tabfirst}0.959 
    \end{tabular}
    }
\end{table}

\begin{table}[htb]
    \centering
    \caption{
        \textbf{Video Depth Estimation: Sintel~\cite{bozic2021transformerfusion}, Bonn~\cite{Palazzolo2019ReFusion3R} and KITTI~\cite{Geiger2013IJRR}.} We have reported results under scale-only alignment in the main paper. Here we further report results using \textbf{joint scale-and-shift} alignment.
    }
    \vspace{-1em}
    
    \resizebox{\linewidth}!{
    
    \begin{tabular}{l c c c c c c c}
        \toprule
        
        \multicolumn{2}{c}{} &
        \multicolumn{2}{c}{\textbf{Sintel}} &
        \multicolumn{2}{c}{\textbf{Bonn}} &
        \multicolumn{2}{c}{\textbf{KITTI}} \\
        
        \cmidrule(lr){3-4} \cmidrule(lr){5-6} \cmidrule(lr){7-8}
        
        \textbf{Method} &
        \textbf{Params} &
        AbsRel$\downarrow$ & $\delta<1.25$$\uparrow$ &
        AbsRel$\downarrow$ & $\delta<1.25$$\uparrow$ &
        AbsRel$\downarrow$ & $\delta<1.25$$\uparrow$ \\
        
        \midrule
        \multicolumn{8}{l}{\textit{$\mathcal{O}(N^2)$ Inference Speed}} \\
        % \midrule
        
        Fast3R~\cite{yang2025fast3r}     & 648M & 0.518 & 0.486 & 0.196 & 0.768 & 0.139 & 0.808 \\

        FLARE~\cite{zhang2025flare}      & 1.40B & 0.791 & 0.358 & 0.142 & 0.797 & 0.357 & 0.579 \\
        VGGT~\cite{wang2025vggt}         & 1.26B & \cellcolor{tabthird}0.226 & \cellcolor{tabthird}0.683 & \cellcolor{tabsecond}0.049 & \cellcolor{tabsecond}0.974 & \cellcolor{tabthird}0.059 & \cellcolor{tabthird}0.961 \\
        $\pi^3$~\cite{wang2025pi3}       & 959M & \cellcolor{tabsecond}0.206 & \cellcolor{tabfirst}0.735 & \cellcolor{tabfirst}0.045 & \cellcolor{tabfirst}0.976 & \cellcolor{tabfirst}0.036 & \cellcolor{tabfirst}0.986 \\

        \midrule
        \multicolumn{8}{l}{\textit{$\mathcal{O}(N)$ Inference Speed}} \\ 

        CUT3R~\cite{wang2025cut3r}       & 793M & 0.534 & 0.551 & 0.067 & 0.961 & 0.124 & 0.850 \\
        TTT3R~\cite{chen2025ttt3r}       & 793M & 0.508 & 0.566 & 0.054 & \cellcolor{tabthird}0.973 & 0.120 & 0.870 \\
        Ours                             & 1.40B & \cellcolor{tabfirst}0.198 & \cellcolor{tabsecond}0.731 & \cellcolor{tabthird}0.052 & \cellcolor{tabthird}0.973 & \cellcolor{tabsecond}0.050 & \cellcolor{tabsecond}0.972 \\
        \bottomrule
    \end{tabular}
    }
    \label{tab:videodepth_scale_shift}
\end{table}

\subsection{Monocular Depth Estimation}
\label{appendix:monodepth_eval}
We present quantitative results for monocular depth estimation in Table~\ref{tab:monodepth}, evaluated on four standard benchmarks. Overall, our method consistently outperforms VGGT~\cite{wang2025vggt} and $\pi^3$~\cite{wang2025pi3}, and performs comparably to the state-of-the-art monocular depth estimator MoGe~\cite{wang2025moge,wang2025mogev2} despite our model never having been trained with purely monocular input.
\subsection{Video Depth Estimation}
We have reported video depth estimation results under scale-only alignment in the main paper. In Table~\ref{tab:videodepth_scale_shift}, we further report results using \textbf{joint scale-and-shift} alignment.

\subsection{Qualitative Comparison}
\label{appendix:qualitative_comparison}

In Figure~\ref{fig:qualitative_comparison}, we show a qualitative comparison on the DTU~\cite{Jensen2014LargeSM} and ETH3D~\cite{schoeps2017cvpr} datasets. Quantitative results for these datasets are shown in Table 3 of the main paper.

\subsection{Effects of Removing the Reference View}
\label{appendix:remove_ref_Vew}

\begin{table}[htb]
    \tablesize
    \centering
    \caption{
        \textbf{Ablation: Removing the Reference View (Camera pose estimation).}
    }
    \vspace{-0.5em}

    \fontsize{6pt}{7pt}\selectfont
    % 2. SET HORIZONTAL PADDING (Tight spacing)
    \setlength{\tabcolsep}{0.8pt}

    \begin{tabular}{l c c c c c c c c c}
        \toprule
        {\multirow{3}{*}{\textbf{Method}}} &
        \multicolumn{3}{c}{\textbf{Sintel}} &
        \multicolumn{3}{c}{\textbf{TUM-dynamics}} &
        \multicolumn{3}{c}{\textbf{ScanNet}~(seen)} \\
        \cmidrule(lr){2-4} \cmidrule(lr){5-7} \cmidrule(lr){8-10}
        &
        ATE$\downarrow$ & RPE trans$\downarrow$ & RPE rot$\downarrow$ &
        ATE$\downarrow$ & RPE trans$\downarrow$ & RPE rot$\downarrow$ &
        ATE$\downarrow$ & RPE trans$\downarrow$ & RPE rot$\downarrow$ \\
        \midrule

        Ours w/ ref                & \textbf{0.125} & \textbf{0.058} & \textbf{0.420} & \textbf{0.012} & \textbf{0.009} & \textbf{0.309} & \textbf{0.034} & \textbf{0.015} & 0.398 \\
        Ours w/o ref                & 0.132 & 0.066 & 0.438 & \textbf{0.012} & 0.010 & 0.310 & \textbf{0.034} & \textbf{0.015} & \textbf{0.385} \\
        \bottomrule
    \end{tabular}
    \label{tab:camera_sintel_tum_scannet_ref_view}

\vspace{1em}

    \centering
    \caption{
        \textbf{Ablation: Removing the Reference View (Video depth estimation).}  We use ``Joint Scale \& Shift'' alignment here.
    }
    \vspace{-0.5em}
    %
    % 1. Font Size & Padding
    \fontsize{6.7pt}{7.4pt}\selectfont
    \setlength{\tabcolsep}{2.5pt} % Tight padding

    \begin{tabular}{l c c c c c c}
        \toprule
         % --- HEADER ROW 1 ---
        % Method and Params are empty here, headers start at col 3
        \multicolumn{1}{c}{} &
         \multicolumn{2}{c}{\textbf{Sintel}} &
        \multicolumn{2}{c}{\textbf{Bonn}} &
         \multicolumn{2}{c}{\textbf{KITTI}} \\
        % cmidrules cover the metric pairs (cols 3-4, 5-6, 7-8)
        \cmidrule(lr){2-3} \cmidrule(lr){4-5} \cmidrule(lr){6-7}
        
         % --- HEADER ROW 2 ---
        \textbf{Method} &
         AbsRel$\downarrow$ & $\delta<1.25$$\uparrow$ &
         AbsRel$\downarrow$ & $\delta<1.25$$\uparrow$ &
         AbsRel$\downarrow$ & $\delta<1.25$$\uparrow$ \\

        \midrule
        Ours w/ ref & 0.205 & \textbf{0.731} & 0.053 & \textbf{0.973} & \textbf{0.048} & \textbf{0.972} \\
         Ours w/o ref & \textbf{0.198} & \textbf{0.731} & \textbf{0.052} & \textbf{0.973} & 0.050 & \textbf{0.972} \\
            \bottomrule
    \end{tabular}
    \label{tab:videodepth_ref_view}

\vspace{1em}

\caption{
\textbf{Ablation: Removing the Reference View (Point Map Estimation).} 
}
\vspace{-1em}

\fontsize{6pt}{7pt}\selectfont

\setlength{\tabcolsep}{1.7pt}

\begin{tabular}{l c c c c c c c c c c c c}
\toprule
% --- HEADER ROW 1 ---
\multirow{3}{*}{\textbf{Method}} &
\multicolumn{6}{c}{\textbf{DTU}} &
\multicolumn{6}{c}{\textbf{ETH3D}} \\
\cmidrule(lr){2-7} \cmidrule(lr){8-13}
% --- HEADER ROW 2 ---
& \multicolumn{2}{c}{Acc. $\downarrow$} &
\multicolumn{2}{c}{Comp. $\downarrow$} &
\multicolumn{2}{c}{N.C. $\uparrow$} &
\multicolumn{2}{c}{Acc. $\downarrow$} &
\multicolumn{2}{c}{Comp. $\downarrow$} &
\multicolumn{2}{c}{N.C. $\uparrow$} \\
\cmidrule(lr){2-3} \cmidrule(lr){4-5} \cmidrule(lr){6-7}
\cmidrule(lr){8-9} \cmidrule(lr){10-11} \cmidrule(lr){12-13}
% --- HEADER ROW 3 ---
& Mean & Med. &
Mean & Med. &
Mean & Med. &
Mean & Med. &
Mean & Med. &
Mean & Med. \\
\midrule
Ours w/ ref & 1.584 & 0.901 & \textbf{1.558} & 0.667 & \textbf{0.687} & \textbf{0.779} & \textbf{0.202} & \textbf{0.138} & 0.413 & 0.278 & 0.852 & 0.953 \\
Ours w/o ref & \textbf{1.228} & \textbf{0.671} & 1.649 & \textbf{0.663} & 0.675 & 0.764 & 0.254 & 0.171 & \textbf{0.249} & \textbf{0.159} & \textbf{0.865} & \textbf{0.965} \\
        \bottomrule
\end{tabular}
\label{tab:pointmap_dtu_eth3d_ref_view}
\end{table}

\begin{figure}[htb]
\centering
\includegraphics[width=\linewidth]{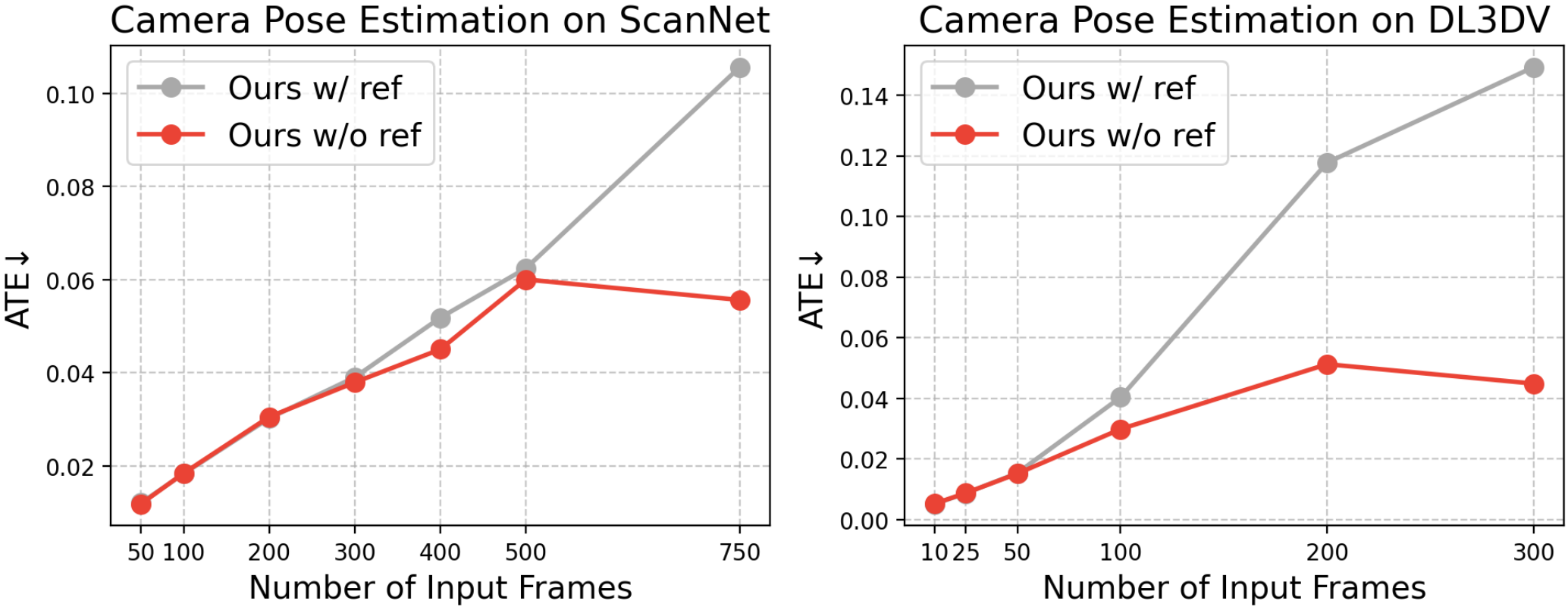}
\caption{
\textbf{Long-sequence camera estimation.} We evaluate camera ATE on the ScanNet-v2 and DL3DV datasets by taking the first $N$ frames of each test sequence and gradually increasing $N$. We see that, when the input sequence length becomes long, removing the reference view and fine-tuning with the affine-invariant camera loss from $\pi^3$~\cite{wang2025pi3} (``Ours w/o ref'') improves the camera pose estimation accuracy compared to the reference-based variant (``Ours w/ ref'').
}
\label{fig:ablation_ref_longseq}
\end{figure}

As described in the main paper, our model is trained in three stages. In the final stage, we remove the explicit reference-view selection: instead of treating the first frame as a reference camera and expressing all poses in its coordinate frame, we fine-tune the model using the affine-invariant camera loss proposed by $\pi^3$~\cite{wang2025pi3}. This loss computes relative pose errors between pairs of views, making the supervision independent of any particular reference frame (see $\pi^3$~\cite{wang2025pi3} for further details).
To evaluate the effect of removing the reference view, we can compare the checkpoint from stage~2 (``Ours w/ ref'') with the checkpoint from stage~3 (``Ours w/o ref''). As shown in Tables~\ref{tab:camera_sintel_tum_scannet_ref_view}, \ref{tab:videodepth_ref_view}, and \ref{tab:pointmap_dtu_eth3d_ref_view}, we find that in our case, neither variant shows a clear or consistent advantage over the other on the standard benchmarks used in Section~4 of the main paper.  That said, we observe that removing the reference view improves accuracy for long input sequences (as shown in Figure~\ref{fig:ablation_ref_longseq}), hence its inclusion in our complete model.

\begin{table}[thb]
    \centering
    \caption{
        \textbf{Streaming Video Depth Estimation.}
        We report results under \textbf{scale-only} alignment on  Sintel~\cite{bozic2021transformerfusion}, Bonn~\cite{Palazzolo2019ReFusion3R} and KITTI~\cite{Geiger2013IJRR}.
    }
    \vspace{-1em}
    
    \fontsize{7pt}{8pt}\selectfont
    \setlength{\tabcolsep}{1pt}
    
    \begin{tabular}{l c c c c c c}
        \toprule
        
        \multicolumn{1}{c}{} &
        \multicolumn{2}{c}{\textbf{Sintel}} &
        \multicolumn{2}{c}{\textbf{Bonn}} &
        \multicolumn{2}{c}{\textbf{KITTI}} \\
        \cmidrule(lr){2-3} \cmidrule(lr){4-5} \cmidrule(lr){6-7}
        
        \textbf{Method} &
        AbsRel$\downarrow$ & $\delta<1.25$$\uparrow$ &
        AbsRel$\downarrow$ & $\delta<1.25$$\uparrow$ &
        AbsRel$\downarrow$ & $\delta<1.25$$\uparrow$ \\
        \midrule
        
        CUT3R~\cite{wang2025cut3r}
            & 0.432 & 0.510
            & 0.072 & 0.951
            & 0.152 & 0.805 \\
        TTT3R~\cite{chen2025ttt3r}
            & \underline{0.426} & \underline{0.522}
            & \textbf{0.061} & \textbf{0.970}
            & \underline{0.149} & \underline{0.812} \\
        \textbf{Ours-streaming}
            & \textbf{0.273} & \textbf{0.638}
            & \underline{0.067} & \underline{0.965}
            & \textbf{0.100} & \textbf{0.903} \\
        
        \bottomrule
    \end{tabular}
    \label{tab:videodepth_streaming_scale_ranked}
\end{table}

\begin{table}[thb]
    \centering
    \caption{
        \textbf{Streaming Camera Pose Estimation: Sintel~\cite{bozic2021transformerfusion} and Co3Dv2~\cite{reizenstein21co3d}.}
        For Sintel, we report ATE and RPE translation/rotation errors.
        For Co3Dv2, we report pose AUC under angular error thresholds of 5/15/30 degrees.   
    }

    \fontsize{7pt}{8pt}\selectfont
    \setlength{\tabcolsep}{1.2pt}
    
    \begin{tabular}{l c c c c c c}
        \toprule
        {\multirow{2}{*}{\textbf{Method}}} &
        \multicolumn{3}{c}{\textbf{Sintel}} &
        \multicolumn{3}{c}{\textbf{Co3Dv2}} \\
        \cmidrule(lr){2-4} \cmidrule(lr){5-7}
        &
        ATE$\downarrow$ & RPE trans$\downarrow$ & RPE rot$\downarrow$ &
        AUC@5$\uparrow$ & AUC@15$\uparrow$ & AUC@30$\uparrow$ \\
        \midrule    
        CUT3R~\cite{wang2025cut3r}   & 0.2160 & \underline{0.0710} & \textbf{0.6220} & \underline{24.88} & \underline{56.28} & \underline{71.72} \\
        TTT3R~\cite{chen2025ttt3r}   & \underline{0.2040} & 0.0850 & \underline{0.6900} & 22.61 & 53.49 & 69.46 \\
        \textbf{Ours-Streaming}      & \textbf{0.1593} & \textbf{0.0655} & 0.7508 & \textbf{45.38} & \textbf{72.58} & \textbf{83.12} \\
        \bottomrule
    \end{tabular}
    \label{tab:streaming_camera_sintel_co3d}
\end{table}

\begin{table}[thb]
    \centering
    \caption{
        \textbf{Streaming Point Map Reconstruction Comparison.}
        We report the results on DTU~\cite{Jensen2014LargeSM}, ETH3D~\cite{schoeps2017cvpr}, and NRGBD-dense.
    }
    \vspace{-0.5em}
    
    \fontsize{6.5pt}{7.5pt}\selectfont
    \setlength{\tabcolsep}{0.4pt}
    
    \begin{tabular}{l c c c c c c c c c}
        \toprule
        
        \multirow{2}{*}{\textbf{Method}} &
        \multicolumn{3}{c}{\textbf{DTU}} &
        \multicolumn{3}{c}{\textbf{ETH3D}} &
        \multicolumn{3}{c}{\textbf{NRGBD-dense}} \\
        \cmidrule(lr){2-4} \cmidrule(lr){5-7} \cmidrule(lr){8-10}
        
        & Acc. $\downarrow$ & Comp. $\downarrow$ & N.C. $\uparrow$ 
        & Acc. $\downarrow$ & Comp. $\downarrow$ & N.C. $\uparrow$
        & Acc. $\downarrow$ & Comp. $\downarrow$ & N.C. $\uparrow$ \\
        \midrule

        CUT3R~\cite{wang2025cut3r}
            & 5.045 & 6.437 & \underline{0.666}
            & \textbf{0.593} & \textbf{0.747} & \textbf{0.754}
            & 0.065 & \underline{0.036} & \underline{0.812} \\
        TTT3R~\cite{chen2025ttt3r}
            & 5.337 & 6.593 & \underline{0.666}
            & 0.763 & \underline{0.881} & 0.739
            & 0.074 & 0.037 & 0.803 \\
        \textbf{Ours-streaming}
            & \textbf{4.091} & \textbf{3.495} & \textbf{0.693}
            & \underline{0.614} & 0.941 & \underline{0.750}
            & \textbf{0.038} & \textbf{0.028} & \textbf{0.836} \\

        \bottomrule
    \end{tabular}
    \label{tab:streaming_recon_comparison_mean}
\end{table}

\subsection{Streaming Reconstruction Comparison}
\label{appendix:streaming_recon}
With a simple fine-tuning procedure, we deploy our model in a streaming setting by updating the TTT-based scene state one view at a time. As shown in Tables~\ref{tab:videodepth_streaming_scale_ranked}, \ref{tab:streaming_recon_comparison_mean}, and \ref{tab:streaming_camera_sintel_co3d},  our streaming variant generally outperforms CUT3R and TTT3R across point-map reconstruction, video depth, and camera pose estimation. Notably, 
due to time constraints we only finetune the model with 24-view context length and our streaming baselines are trained with longer contexts (up to 64 views), we reasonably expect that further scaling the finetuning context length will yield an even larger advantage.

\subsection{More Long-Sequence Evaluation}
\label{appendix:more_long_eval}
\begin{figure}[htp]
\centering
\includegraphics[width=0.9\linewidth]{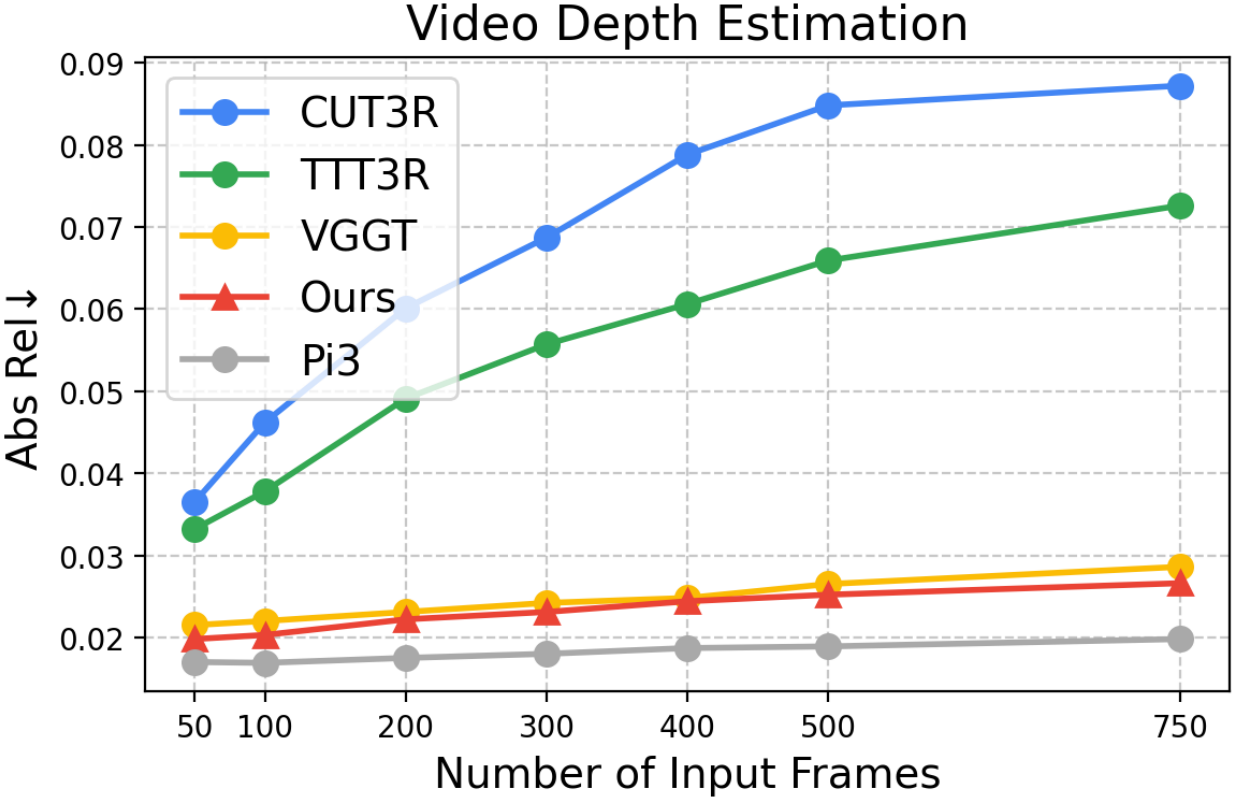}
\caption{
\textbf{Long-sequence video depth estimation.} We evaluate on the ScanNet-v2 dataset by taking the first $N$ frames of each test sequence and gradually increasing $N$.
}
\label{fig:longseq_eval_videodepth}
\end{figure}

\begin{figure}[htp]
\centering
\includegraphics[width=1.02\linewidth]{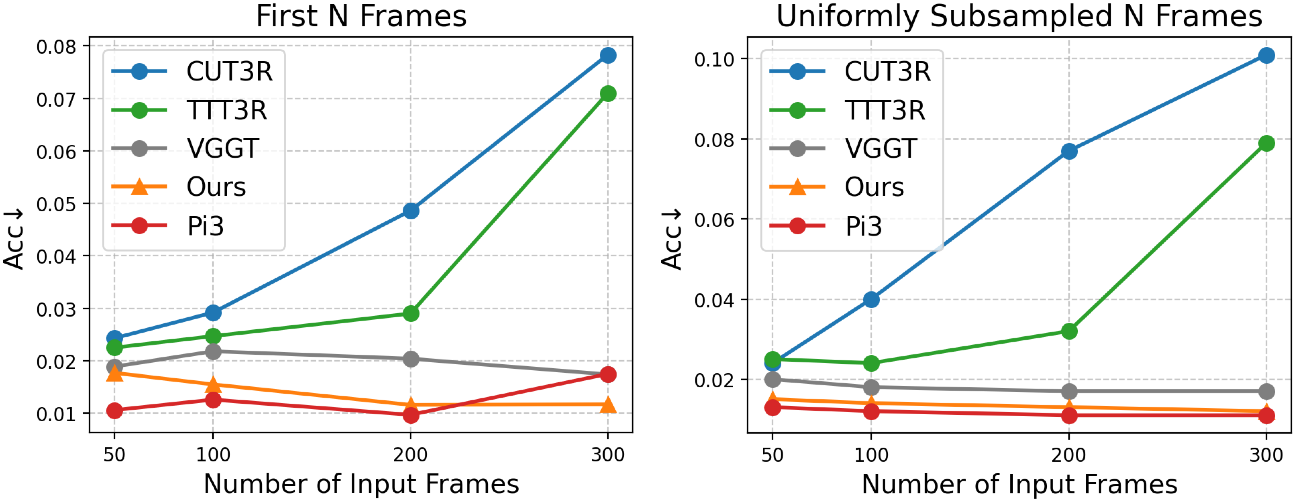}
\caption{
\textbf{Long-sequence 3D point estimation.} We evaluate on the 7-Scenes~\cite{Shotton2013SceneCR} dataset under two cases. \textbf{Left}: increasing \emph{scene scale} by using the first $N$ frames of each sequence; \textbf{Right}: increasing \emph{view density} by uniformly subsampling $N$ frames.
}
\label{fig:longseq_eval_7scenes}
\end{figure}

We show more long sequence evaluation results on video depth estimation and 3D point estimation in Figure~\ref{fig:longseq_eval_videodepth} and Figure~\ref{fig:longseq_eval_7scenes}, respectively.

\begin{table}[thb]
    \centering
    \caption{
        \textbf{NVS Evaluation on Mip-NeRF360~\cite{barron2022mipnerf360}.}
        Baseline results are taken from Tab.~1 of the AnySplat paper. ZipMap is competitive with several baselines. However, as noted in our Limitations, our query results are primarily designed to generate colored point clouds rather than high-fidelity novel view synthesis. Accordingly, NVS is not our main focus, and we do not claim state-of-the-art performance compared with explicit 3DGS-based methods.
    }
    \label{tab:nvs_mipnerf360}

    \resizebox{\linewidth}{!}{
        \begin{tabular}{l ccc ccc}
            \toprule
            \multirow{2}{*}{Method} & \multicolumn{3}{c}{6 Views} & \multicolumn{3}{c}{16 Views} \\
            \cmidrule(lr){2-4} \cmidrule(lr){5-7}
            & PSNR $\uparrow$ & SSIM $\uparrow$ & LPIPS $\downarrow$
            & PSNR $\uparrow$ & SSIM $\uparrow$ & LPIPS $\downarrow$ \\
            \midrule
            NoPoSplat~\cite{ye2024noposplat} & 15.92 & 0.416 & 0.541 & 15.47 & 0.361 & 0.606 \\
            Flare~\cite{zhang2025flare}     & 15.35 & 0.407 & 0.573 & 13.21 & 0.348 & 0.695 \\
            AnySplat~\cite{jiang2025anysplat}  & 18.32 & 0.524 & 0.329 & 21.85 & 0.670 & 0.250 \\
            Ours & 15.60 & 0.414 & 0.486 & 17.65 & 0.459 & 0.374 \\
            \bottomrule
        \end{tabular}
    }
\end{table}

\section{Limitations}
\label{appendix:limitations}
Though our model achieves high reconstruction accuracy and fast inference speeds, it has several limitations.
First, we observe noticeable performance degradation when evaluating very long sequences where the scene scale extends far beyond the training distribution. This limitation appears to be shared by all existing feed-forward methods. Promising directions to address this issue include: (i) training the model on longer sequences using efficient large-context training strategies such as context parallelism (CP); and (ii) combining our methods with global alignment techniques, such as VGGT-Long~\cite{deng2025vggtlongchunkitloop}. Thankfully, because our method offers significantly faster runtimes on long sequences compared to prior approaches, it may be well-suited to training on longer sequences at higher throughput rates than prior models.
Second, although our model can query novel-view RGB information from the implicit scene representation, the rendered novel views often exhibit blurry artifacts in high-frequency regions. We suspect this is partly due to a mismatch  between the GT pose condition and the implicit scene geometry learned from unposed images, which leads to noisy RGB supervision during training.  As shown in Tab~\ref{tab:nvs_mipnerf360}, we do not claim that our current method enables SoTA unposed novel view synthesis, and our query results are primarily designed to generate colored point clouds. Improving the RGB rendering quality of our model to support high-fidelity unposed novel view synthesis remains an interesting future work.

% \clearpage
% \FloatBarrier
% {\small
% \bibliographystyle{ieeenat_fullname}
% \bibliography{main}
% }